\documentclass[10pt,journal,compsoc]{IEEEtran}
\ifCLASSOPTIONcompsoc
  \usepackage[nocompress]{cite}
\else
  \usepackage{cite}
\fi

\ifCLASSINFOpdf
\else
\fi
\usepackage{sidecap}  %required for side captions
\usepackage{caption}
\usepackage{subcaption}

\usepackage[capbesideposition=outside,capbesidesep=quad]{floatrow}

\usepackage[utf8]{inputenc} % allow utf-8 input
\usepackage[T1]{fontenc}    % use 8-bit T1 fonts
\usepackage{hyperref}       % hyperlinks
\usepackage{url}            % simple URL typesetting
\usepackage{booktabs}       % professional-quality tables
\usepackage{amsfonts}       % blackboard math symbols
\usepackage{nicefrac}       % compact symbols for 1/2, etc.
\usepackage{microtype}      % microtypography
\usepackage{epsfig}
\usepackage{graphicx}
\usepackage{amsmath}
\usepackage{amssymb}
\usepackage{multirow}
\usepackage[capbesideposition=outside,capbesidesep=quad]{floatrow}
\usepackage{lipsum}    % Dummytext

\usepackage{listings}
\usepackage{latexsym}
\usepackage{algpseudocode}
\usepackage{sidecap}
\usepackage{blindtext}
\usepackage{algorithm}
\usepackage{textcase}
\usepackage{comment}
\usepackage{breqn}
\usepackage{ragged2e}
\usepackage[normalem]{ulem}
\usepackage{caption, booktabs}
\usepackage{caption}
\usepackage{bbm}

\usepackage{xparse}
\usepackage{amssymb}

\usepackage{array}
\newcolumntype{L}[1]{>{\raggedright\let\newline\\\arraybackslash\hspace{0pt}}m{#1}}
\newcolumntype{C}[1]{>{\centering\let\newline\\\arraybackslash\hspace{0pt}}m{#1}}
\newcolumntype{R}[1]{>{\raggedleft\let\newline\\\arraybackslash\hspace{0pt}}m{#1}}
\newcolumntype{J}[1]{>{\justifying\let\newline\\\arraybackslash\hspace{0pt}}m{#1}}

\usepackage{enumitem}
\usepackage{float}

\newcommand{\myComment}[1]{}          % comment

\usepackage[table,xcdraw]{xcolor}
\usepackage[colorinlistoftodos,prependcaption,textsize=tiny]{todonotes}

\definecolor{orange}{rgb}{1,0.6,0}
\definecolor{midNightBlue}{rgb}{0.01,0.01, 0.55}
\begin{document}

\onecolumn

\section*{IEEE copyright notice}

© 2023 IEEE.  Personal use of this material is permitted.  Permission from IEEE must be obtained for all other uses, in any current or future media, including reprinting/republishing this material for advertising or promotional purposes, creating new collective works, for resale or redistribution to servers or lists, or reuse of any copyrighted component of this work in other works.

\section*{Cite as}

M. A. Munir, M. H. Khan, M. S. Sarfraz and M. Ali, "Domain Adaptive Object Detection via Balancing Between Self-Training and Adversarial Learning," in IEEE Transactions on Pattern Analysis and Machine Intelligence, vol. 45, no. 12, pp. 14353-14365, Dec. 2023, doi: 10.1109/TPAMI.2023.3290135.

\section*{Bibtex}

\begin{lstlisting}
@ARTICLE{domain2023munir,
  author={Munir, Muhammad Akhtar and Khan, Muhammad Haris and Sarfraz, M. Saquib 
        and Ali, Mohsen},
  journal={IEEE Transactions on Pattern Analysis and Machine Intelligence}, 
  title={Domain Adaptive Object Detection via Balancing Between Self-Training 
        and Adversarial Learning},  
  year={2023},
  volume={45},
  number={12},
  pages={14353-14365},
  doi={10.1109/TPAMI.2023.3290135}}
\end{lstlisting}

\section*{Final published article}

https://ieeexplore.ieee.org/document/10173487

\newpage
\twocolumn
\title{Domain Adaptive Object Detection via Balancing between Self-Training and Adversarial Learning}

\author{Muhammad~Akhtar~Munir,
        Muhammad~Haris~Khan,
        M.~Saquib~Sarfraz,
        and~Mohsen~Ali% <-this % stops a space
\IEEEcompsocitemizethanks{\IEEEcompsocthanksitem M.A Munir and M. Ali are with the Department
of Computer Science, Information Technology University of Punjab, Pakistan.\protect\\

E-mail: \{akhtar.munir, mohsen.ali\}@itu.edu.pk
\IEEEcompsocthanksitem M. H. Khan and M. S. Sarfraz are with Mohamed bin Zayed University of Artificial Intelligence and Karlsruhe Institute of Technology.}% <-this % stops a space
\thanks{Manuscript received ; revised.}}

% \markboth{IEEE TPAMI,~Vol.~-, No.~-, March~2022}%
% {Akhtar \MakeLowercase{\textit{et al.}}: Domain Adaptive Object Detection via Balancing between Self-Training and Adversarial Learning}

\IEEEtitleabstractindextext{%

\begin{abstract}
Deep learning based object detectors struggle generalizing to a new target domain bearing significant variations in object and background. Most current methods align domains by using image or instance-level adversarial feature alignment. This often suffers due to unwanted background and lacks class-specific alignment. A straightforward approach to promote class-level alignment is to use high confidence predictions on unlabeled domain as pseudo-labels. These predictions are often noisy since model is poorly calibrated under domain shift. 
In this paper, we propose to leverage model’s predictive uncertainty to strike the right balance between adversarial feature alignment and class-level alignment. We develop a technique to quantify predictive uncertainty on class assignments and bounding-box predictions. Model predictions with low uncertainty are used to generate pseudo-labels for self-training, whereas the ones with higher uncertainty are used to generate tiles for adversarial feature alignment. This synergy between tiling around uncertain object regions and generating pseudo-labels from highly certain object regions allows capturing both image and instance-level context during the model adaptation.
We report thorough ablation study to reveal the impact of different components in our approach. Results on five diverse and challenging adaptation scenarios show that our approach outperforms existing state-of-the-art methods with noticeable margins.

\end{abstract}

\begin{IEEEkeywords}
Unsupervised Domain Adaptation, Uncertainty, Object Detection, Self Training, Adversarial Learning.
\end{IEEEkeywords}}

\maketitle

\IEEEdisplaynontitleabstractindextext

\IEEEpeerreviewmaketitle

\ifCLASSOPTIONcompsoc
\IEEEraisesectionheading{\section{Introduction}\label{sec:introduction}}
\else
\section{Introduction}
\label{sec:introduction}
\fi

\IEEEPARstart{W}{e} have seen remarkable progress in convolutional neural network based object detectors, owing to their capability of learning representative features from large annotated datasets \cite{everingham2010pascal, lin2014microsoft, kitty2012we}.
However, akin to other supervised deep learning methods, object detectors trained on the source domain struggle generalizing adequately to a new target domain.
This is a well-known \textit{domain shift} problem \cite{torralba2011unbiased}, typically caused by change in style, camera pose, or object size and orientation, or the number or location of objects in the scene, among other things.
Often, collecting large annotated dataset for supervised adaptation to the target domain is expensive, error prone and in many cases not possible. 
Unsupervised Domain Adaptation (UDA) is a promising research direction for solving this problem by transferring knowledge from a labelled source domain to an unlabelled target domain.

Many unsupervised domain adaptive detectors rely on adversarial adaptation or self-training techniques.
Methods based on adversarial adaptation \cite{chen2018domain,saito2019strong, he2019multi, hsu2020every, zheng2020cross, xu2020exploring, chen2020harmonizing, nguyen2020uada, seeking_2021_ICCV}, mostly use domain discriminator for aligning features at image or instance level.  
However, due to the absence of ground truth annotations in the target domain they suffer from the challenges of how to select samples for the adaptation. 
Uniform selection is a straightforward approach, however, it is prone to missing on infrequent classes or instances.
Most importantly adversarial alignment do not explicitly incorporates class discriminative information, and could result in non-optimal alignment for classification and object detection tasks \cite{saito2019strong, chen2018domain, 2018dirt}.
A potential solution to this problem is self-training based adaptation, however, it faces the challenge of how to avoid noisy pseudo-labels.
Some methods choose high confidence predictions as pseudo-labels  \cite{lee2013pseudo, inoue2018cross, roychowdhury2019automatic}, but the likely poor calibration of model under domain shift renders this solution inefficient \cite{NEURIPS2019_canyou}.
Further, in the case of object detection, prediction probability can not directly capture object localization inaccuracies.

We present a principled approach, coined as SSAL (Synergizing between Self-Training and Adversarial Learning for Domain Adaptive Object Detection)%
, to achieve the right balance between self-training and adversarial alignment for domain adaptive object detection via leveraging model's predictive uncertainty. 
To estimate predictive uncertainty of a detection, we propose taking into account variations in both the localization prediction and confidence prediction across Monte-Carlo dropout inferences \cite{gal2016dropout}. 
Certain detections are taken as pseudo-labels for self-training, while uncertain ones are used to extract tiles (regions in image) for adversarial feature alignment.
This synergy between adversarial alignment via tiling around the uncertain object regions and self-training with pseudo-labels from certain object regions allow us include instance-level context for effective adversarial alignment and improve feature discriminability for class-specific alignment.
Since we select pseudo-labels with low uncertainty for self-training and take relatively uncertain as potential, object-like regions with context (i.e. tiles) for adversarial alignment, we tend to reduce the effect of poor calibration under domain shift, thereby improving model's generalization across domains. 

We summarize our key contributions as follows. We introduce a new uncertainty-guided framework that strikes the right balance between self-training and adversarial feature alignment for adapting object detection methods. Both pseudo-labelling for self-training and tiling for adversarial alignment are impactful due to their simplicity, generality and ease of implementation. We propose a method for estimating the object detection uncertainty via taking into account variations in both the localization prediction and confidence prediction across Monte-Carlo dropout inferences. We show that, selecting pseudo-labels with low uncertainty and using relatively uncertain regions for adversarial alignment, it is possible to address the poor calibration under domain shift, and hence improve model's generalization across domains. Unlike most of the previous methods, we build on computationally efficient one-stage anchor-less object detectors and achieve state-of-the-art results with notable margins across various adaptation scenarios.

    A preliminary version of this work appeared in \cite{munir2021ssal}. In addition, the current manuscript makes following new contributions. We revisit the uncertainty quantification mechanism for object detection to incorporate a new constraint for selecting pseudo-labels. 
    \textcolor{black}{The tile set is extended via relaxing the uncertainty-guided tiling constraint and including randomly sampled full image.}
    % \sout{We extend the tile set via relaxing the uncertainty-guided tiling constraint and including randomly sampled full image.}
    After revisiting uncertainty quantification, incorporating new constraint, and extending the tiling set, we dub our framework as SSAL\textsuperscript{\textdagger}. We include extensive ablation studies to \textcolor{black}{analyze} SSAL\textsuperscript{\textdagger} and draw comparisons with its previously published conference version SSAL \cite{munir2021ssal}.
Finally, we include experimental results on two new large-scale and challenging adaptation scenarios, encompassing severe domain shifts.

\begin{figure*}[t]
% \begin{center}
\centering
\includegraphics[width=0.7\linewidth]{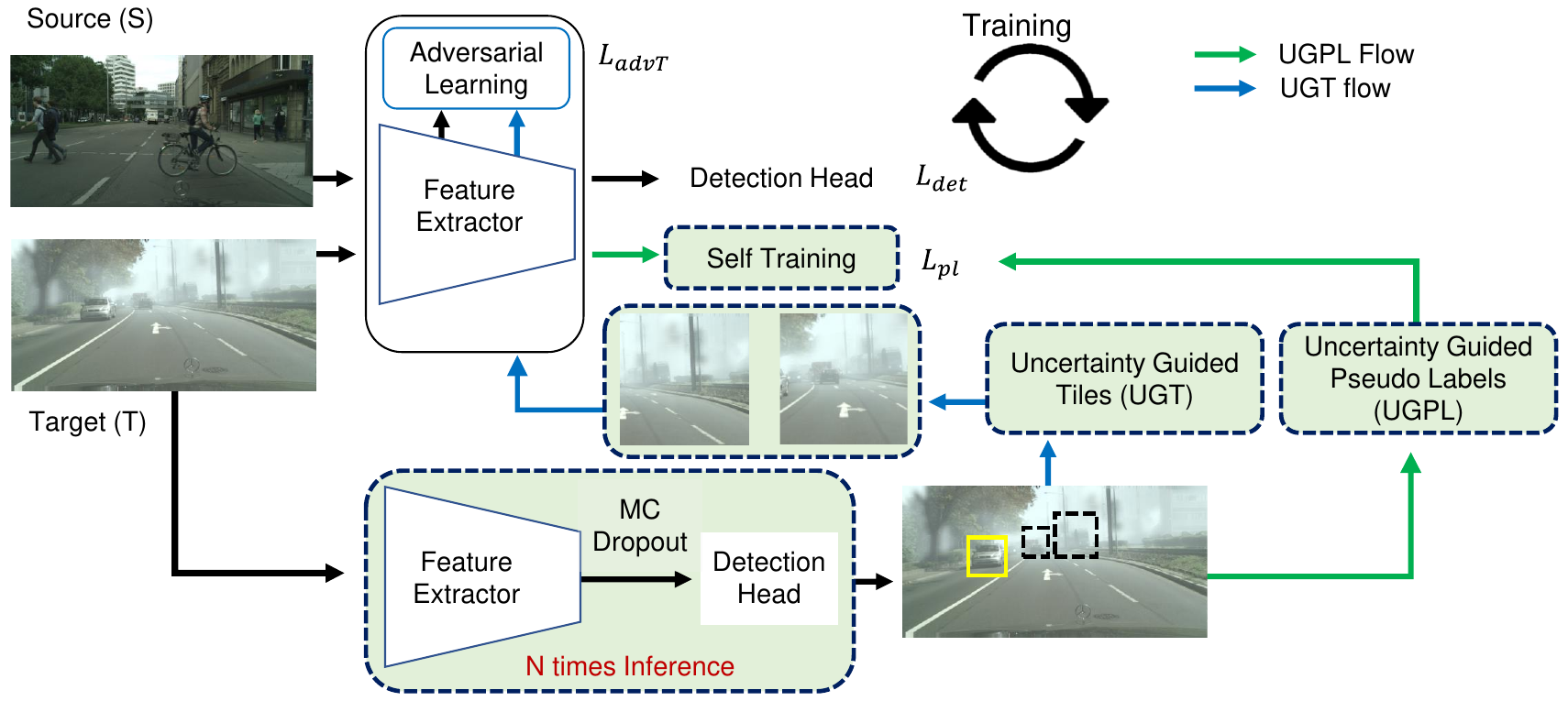}
\captionsetup{justification=justified}
\caption{Overall architecture of SSAL. Fundamentally, it is a one-stage detector \cite{tian2019fcos} with an adversarial feature alignment stage. We propose uncertainty-guided pseudo-labelling (UGPL) for self-training and uncertainty-guided tiling (UGT) for adversarial alignment (dotted boxes). UGPL produces accurate pseudo-labels in target image which are used with ground-truth labels in source image for training. UGT extracts tiles around possibly object-like regions in target image which are used with randomly extracted tiles around ground-truth labels in source domain for adversarial feature alignment.\vspace{-0.7em}
}
%\vspace{-0.4cm}
\label{fig:architecture_diagram}
% \end{center}
\end{figure*}

\section{Related Work}
Domain adaptation has been studied in various computer vision tasks including classification\cite{prabhu2020sentry, liu2021cycle}, semantic segmentation \cite{michieli2020adversarial, subhani2020learning, iqbal2022fogadapt, vu2019advent, yang2020fda}, and object detection\cite{he2019multi, kim2019self, xu2020cross, chen2020harmonizing, jiang2021decoupled}. Below, for brevity, we only present an extensive literature survey on object detection and its domain adaptation application.

\noindent \textbf{Object detection.} Deep learning based object detection algorithms can be broadly classified into either anchor-based \cite{DBLP:conf/nips/RenHGS15,DBLP:conf/cvpr/LinDGHHB17, DBLP:conf/cvpr/SinghD18,DBLP:conf/cvpr/CaiV18} or anchor-free methods \cite{law2018cornernet,duan2019centernet,tian2019fcos}. Anchor-based methods, such as Faster-RCNN \cite{DBLP:conf/nips/RenHGS15}, uses region proposal network (RPN) to generate proposals.
\textcolor{black}{RPN is trained with a subsequent stage, for bounding box regression and classfication, in end-to-end fashion to classify region of interests, thus making it a two-stage object detector}. 
Anchor-free detectors, on the other hand, skip proposal generation step and through leveraging fully convolutional network (FCN) \cite{long2015fully} directly localize objects. For instance, \cite{tian2019fcos} proposed per-pixel prediction and directly predicted the class and offset of the corresponding object at each location on the feature map. In this work, we capitalize on the computationally inexpensive characteristic in anchor-free detectors to study adapting trained object detectors.

\noindent \textbf{Tiling for object detection.} The process of cropping regions of an input image, a.k.a tiling, in a uniform \cite{ozge2019power}, random, or informed \cite{yang2019clustered,hong2019patch,li2020density} fashion before detection pipeline is typically used to tackle scale variation problem and improve detection accuracy over small objects. 
Informed tiling can be achieved by first generating a set of regions of object clusters, and then cropping them for subsequent fine detection \cite{yang2019clustered}.

\noindent \textbf{Domain-adaptive object detection.}  The pioneering work of \cite{chen2018domain} on domain-adaptive (DA) object detection proposed reducing domain shift at both image and instance levels via embedding adversarial feature adaptation into anchor-based detection pipeline. Global feature alignment could suffer as domains may manifest distinct scene layouts and complex object combinations. Several subsequent approaches attempted to achieve a right balance between the global and instance-level alignments \cite{zhu2019adapting,xu2020exploring}. Other methods \cite{he2019multi, kim2019diversify, cai2019exploring, hsu2020progressive} improved feature alignment in various ways e.g., through exploiting hierarchical feature learning in CNNs \cite{he2019multi}.
\cite{sultani2022towards}
explored  partially supervised domain adaptation for object detection. 
\textcolor{black}{Recently, \cite{seeking_2021_ICCV} employed clustering to group visually similar proposals and performed adversarial alignment on image-level and group-level cluster features.
}

While above methods are built on two-stage pipeline, a few approaches have built domain adaptive detectors on one-stage pipeline \cite{kim2019self, hsu2020every}. \cite{hsu2020every} proposed to predict pixel-wise objectness and center-aware feature alignment, building on \cite{tian2019fcos}, to focus on the discriminative parts of objects. 
\textcolor{black}{Using one stage domain adaptive approach, \cite{mining_2021_ICCV} first aligned features using foreground/background classifier across the domains, and further alignment is based on categorical consistency across the domains.}

\noindent \textbf{Uncertainty for DA object detection.} Exploiting model's predictive uncertainty and entropy optimization have remained subject of interest in prior cross-domain recognition \cite{long2018conditional, han2019unsupervised, manders2018adversarial, ringwald2020unsupervised} and detection  \cite{guan2021uncertainty,nguyen2020domain} works. 
For cross-domain recognition, \cite{ringwald2020unsupervised} employed uncertainty for filtering training data and aligning features in Euclidean space. 
For DA object detection, \cite{guan2021uncertainty} proposed an uncertainty metric to regulate the strength of adversarial learning for well-aligned and poorly-aligned samples adaptively. 

\noindent \textbf{Pseudo-labelling for DA object detection.} %Pseudo-labelling, used for model self-training, generates pseudo-labels for unlabelled samples with a model trained on labelled data. 
In DA object detection, pseudo-labelling aims at acquiring pseudo instance-level annotations for incorporating discriminative information. \cite{inoue2018cross} generated pseudo instance-level annotations by choosing the top-1 confidence detections.
Similarly, \cite{roychowdhury2019automatic} obtained the same by using high-confidence detections and further refined them using tracker's output. 
Towards refining (noisy) pseudo instance-level annotations, \cite{khodabandeh2019robust} employed auxiliary component and \cite{kim2019self} devised a criterion based on supporting RoIs. 
Confidence-based pseudo-label selection is prone to generating noisy labels since the model is poorly calibrated under domain shift, eventually causing degenerate network re-training. 

Unlike most prior methods we build on computationally inexpensive one-stage anchor-free detector. Different to existing methods, we leverage model's predictive uncertainty, considering variations in localization and confidence predictions across MC simulations, to achieve the best of both self-training and adversarial alignment through mining highly certain target detections as pseudo-labels and relatively uncertain ones as guides in the tiling process.

\section{Overall Framework}

%In this section, we describe the technical details of our method. 
Fig.~\ref{fig:architecture_diagram} displays the overall architecture of our method. We propose to leverage model's predictive uncertainty to strike the right balance between adversarial feature alignment and self-training. To this end, we introduce uncertainty-guided pseudo-labels selection (UGPL) for self-training and uncertainty-guided tiling (UGT) for adversarial alignment. The former allows generating accurate pseudo-labels to improve feature discriminability for class-specific alignment, while the latter enables extracting tiles on uncertain, object-like regions for effective domain alignment.

\subsection{Preliminaries}

\noindent \textbf{Problem Setting.} 
Let $\mathcal{D}_{s} = \{(x_{i}^{s}, \mathbf{y}_{i}^{s})\}_{i=1 }^{N_s}$ be the labeled source dataset and $\mathcal{D}_{t} = \{x_{j}^{t}\}_{j=1 }^{N_t}$ be the unlabeled target dataset. Where $\mathbf{y}_{i}^{s} =\{\mathbf{b}_i^s, \mathbf{c}_i^s\}$ is set of bounding boxes $\mathbf{b}_i^s$ for the objects in the image  $x_{i}^{s}$ and their corresponding classes $\mathbf{c}_i^s \in \{1, \dots, C\}$. The source and target domains share an identical label space, however, violate the i.i.d. assumption since they are sampled from different data distributions. Our goal is to learn a domain-adaptive object detector, given labeled $\mathcal{D}_{s}$ and unlabeled $\mathcal{D}_{t}$, capable of performing accurately in the target domain.

\noindent \textbf{One-stage anchor-free object detection.} Owing to the computationally inexpensive feature of one-stage anchor-free detection pipelines, we build our uncertainty-guided domain-adaptive detector on fully convolutional one-stage object detector (FCOS) \cite{tian2019fcos}. Inspired from the fully convolutional architecture \cite{long2015fully}, FCOS incorporates per-pixel predictions and directly regresses object location. Specifically, it outputs a $C$-dimensional classification vector, a 4D vector of bounding box coordinates, and a centerness score. The loss function for training FCOS is: 
\begingroup
\normalsize
\begin{flalign}
\begin{split}
    \mathcal{L}_{det}(\mathbf{c}_{u,v}, \mathbf{b}_{u,v}) = \frac{1}{N_{pos}} \sum_{u,v}  \mathcal{L}_{cls}(\widehat{\mathbf{c}}_{u,v}, c_{u,v}) \\
    + \frac{1}{N_{pos}} \sum_{u,v} \mathbbm{1}_{\widehat{c}_{u,v} > 0} \mathcal{L}
    _{box}(\widehat{\mathbf{b}}_{u,v}, \mathbf{b}_{u,v}) 
    \label{eq:det}
\end{split}
% \label{eq:fcosLoss}
\end{flalign}
\endgroup

where $\mathcal{L}_{cls}$ is the classification loss (i.e. focal loss \cite{lin2018focal}, and $\mathcal{L}_{box}$ (i.e. IoU loss \cite{yu2016unitbox}) is the regression loss. $\widehat{\mathbf{c}}_{u,v}, \widehat{\mathbf{b}}_{u,v}$ denotes class and bounding box predictions at location $(u,v)$. $N_{pos}$ denotes the number of positive samples. 

\noindent \textbf{Adversarial feature alignment.} Several methods \cite{saito2019strong, chen2018domain} align feature maps on the image-level to reduce domain shift via adversarial learning. It involves a global discriminator $D_{adv}$ that identifies whether the pixels on each feature map belong to the source or the target domain. Specifically, let $F \in \mathbb{R}^{H\times W \times K}$ be the $K$-dimensional feature map of spatial resolution $H \times W$ extracted from the feature backbone network. The output of $D_{adv}$ is a domain classification map of the same size as $F$. The discriminator can be optimized using binary cross-entropy loss:
\begingroup
\small
\begin{flalign}
\begin{split}
    \mathcal{L}_{adv}(x^s, x^t) = -\sum_{u,v} q \log(D_{adv}(F^s)_{u,v}), \\
    + (1-q) \log(1-D_{adv}(F^t)_{u,v})
    \label{eq:ugte}
\end{split}
\end{flalign}
\endgroup

where $q$ is the domain label $\in \{0,1\}$. We perform adversarial feature alignment by applying gradient reversal layer (GRL) \cite{ganin2015unsupervised} to source $F^s$ and target $F^t$ feature maps, in which the sign of gradient is flipped when optimizing the feature extractor via GRL layer. Global alignment is prone to focusing on (unwanted) background pixels. We introduce uncertainty-guided tiling, that involves cropping tiles (regions with context) around object-like regions for effective adversarial alignment (sec.~\ref{sec:SSAL}).

\begin{figure*}[t]
% \begin{center}
        \includegraphics[width=1.0\linewidth]{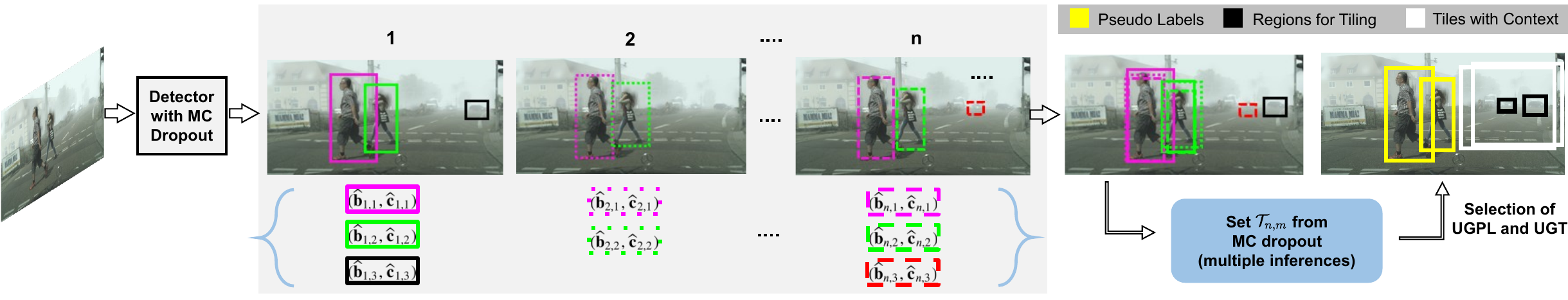}\captionsetup{justification=justified}
        \caption{\small An illustration showing the detections that are either taken as pseudo-labels or used to extract tiles. More certain detections, such as pedestrians are taken as pseudo-labels, whereas relatively uncertain ones, like cars under fog, are used for extracting tiles.
        Eq.~\ref{eq:tset} constructs the set from the MC dropout mechanism. Using Eqs.~\ref{eq:uncer} \& \ref{eq:uncersd}, our method quantifies the uncertainty and based on this information selection of UGPL and UGT takes place (Eqs.~\ref{eq:plSelec_3} \& \ref{eq:tile_1})
        }        \label{fig:illustration_certain_vs_uncertain}
% \end{center}
\end{figure*}

\noindent \textbf{Self-Training.} Self-training is a process of training with pseudo-labels, which are generated for unlabelled samples in the target domain with a model trained on labelled data. Hard pseudo instance-level labels are obtained directly from network class predictions. Let $\mathbf{p}_{j,k}$ be the probability outputs vector of a trained network corresponding to a detection $\widehat{\mathbf{y}}_{j,k}$, such that $p_{j,k}^c$ denotes the probability of class $c$ being present in the detection. With these probabilities, the pseudo-label can be generated for $\widehat{\mathbf{y}}_{j,k}$  as: $\tilde{y}_{j,k}^{c} = \mathbbm{1}[p_{j,k}^c\geq \alpha]$, where $\alpha = max_{c}p_{j,k}^c$. There could be a significant fraction of incorrectly pseudo-labelled detections used during training. A common strategy to reduce noise during training is to select pseudo-labels corresponding to high-confidence detections \cite{inoue2018cross,roychowdhury2019automatic}. Let $g_{j,k}$ be a boolean variable denoting the selection or rejection of $\tilde{y}_{j,k}$ i.e. where $g_{j,k}=1$ when $\tilde{y}_{j,k}$ is selected or otherwise. Formally, in confidence-based selection, a pseudo-label $\tilde{y}_{j,k}$ is selected as: $g_{j,k}=\mathbbm{1}[p_{j,k}^c\geq \tau]$, where $\tau$ is the confidence threshold. These high confidence detections are often noisy because the model is poorly calibrated under domain shift. Instead, we propose to select pseudo-labels utilizing uncertainty in both class prediction and localization prediction to mitigate the impact of poor network calibration (sec.~\ref{sec:SSAL}).

\vspace{-0.6em}
\subsection{SSAL}
\label{sec:SSAL}
The source model demonstrates poor calibration under target domain bearing sufficiently different superficial statistics and different object combinations \cite{NEURIPS2019_canyou, 2018dirt}. Although confidence-based selection (typically highest confidence) improves accuracy, the poor calibration of the model under domain shift makes this strategy inefficient. As a result, it could lead to both poor pseudo-labelling accuracy and incorrect identification of possibly object-like regions for adversarial alignment. Since calibration can be considered as the model's overall prediction uncertainty \cite{lakshminarayanan2016simple}, we believe that through leveraging model's predictive uncertainty we can negate the poor effects of calibration. To this end, we propose to leverage uncertainty in detections to select pseudo-labels for self-training and choose regions for tiling in adversarial alignment (see Fig.~\ref{fig:illustration_certain_vs_uncertain}).

\noindent \textbf{Uncertainty in object detections.} Assuming one stage detector, we estimate detection uncertainty by applying Monte-Carlo dropout \cite{gal2016dropout} (in particular, spatial dropout \cite{tompson2015efficient}) to the convolutional filters after the feature extraction layer. Given an image $x$, we perform $N$ stochastic forward passes (inferences) using MC dropout. Let $\widehat{\mathbf{y}}_{n,m} = (\widehat{\mathbf{b}}_{n,m}, \widehat{c}_{n,m}) $ be the $m_{th}$ detection in $n_{th}$ inference, $\widehat{c}_{n,m}$ be the class label with highest probability $\widehat{p}_{n,m}$ in the probability vector $\mathbf{p}_{n,m}$, and $\widehat{\mathbf{b}}_{n,m} \in \mathbb{R}^{4}$ is the predicted bounding box. 
We aim to capture the variations in both the localization prediction and confidence prediction across inferences. To this end, we define the uncertainty of the object detection prediction as the mean class probability of the overlapping bounding boxes across individual inferences.

Specifically, for each $\widehat{\mathbf{y}}_{n,m}$, we create a set $\mathcal{T}_{n,m}$ by including all $\widehat{\mathbf{y}}_{k,l}$, where $k \ne n$ and $l$ is an arbitrary detection in $k_{th}$ MC forward pass, such that $\widehat{\mathbf{b}}_{n,m}$ has IoU with $\widehat{\mathbf{b}}_{k,l}$ greater than a specific threshold and $\widehat{c}_{n,m}=\widehat{c}_{k,l}$.
% \begin{equation}
\begingroup
\small
\begin{flalign}
\begin{split}
    \mathcal{T}_{n,m}= \{ \forall_{k \ne n} \cup (\widehat{\mathbf{b}}_{k,l} , \widehat{c}_{k,l}),~|~IoU(\widehat{\mathbf{b}}_{n,m},\widehat{\mathbf{b}}_{k,l})>\gamma  ~, ~  \\ \widehat{c}_{k,l} = \widehat{c}_{n,m} ~ \}
    \label{eq:tset}
\end{split}
\end{flalign}
\endgroup
% \end{equation}

Where $\gamma$ is the IoU threshold to identify bounding boxes occupying same region (detected as same object). We use $\mathcal{T}_{n,m}$ to estimate uncertainty based on both localization prediction and confidence prediction for $\widehat{\mathbf{y}}_{n,m}$ as:
\begin{equation}
    \label{eq:uncer}
     \hat{p}_{n,m} = \frac{1}{|\mathcal{T}_{n,m}|} \sum_{e} \widehat{p}_{n,m}^{e},
\end{equation}

where $\widehat{p}_{n,m}^{e}$ is the class prediction confidence of $e_{th}$ detection in $\mathcal{T}_{n,m}$. See Fig.~\ref{fig:illustration_certain_vs_uncertain} for an illustration of quantifying detection uncertainty.

We interpret the averaged confidences $\hat{p}_{(.)}$
as a proxy (or indirect) measure of how uncertain (or certain) the model is in its class assignment and object localization \cite{ringwald2020unsupervised}. Under this definition, the model will be completely uncertain if $\hat{p}_{(.)}$ has uniform distribution whereas it will be completely certain if $\hat{p}_{(.)}$ can be represented by a Kronecker delta function.

\begin{figure}[t]
    \centering
    \includegraphics[width=1.0\linewidth]{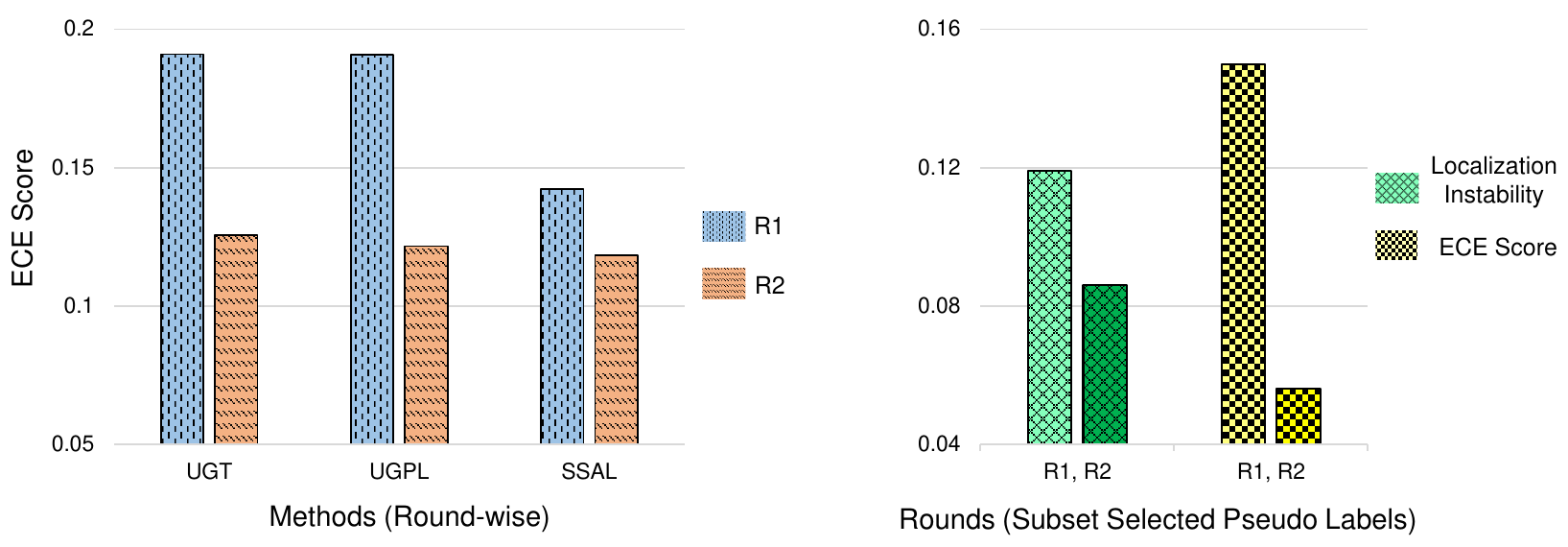}
    %\vspace{-0.3cm}
    \captionsetup{justification=justified}
    \caption{\small 
    \textbf{Left.} ECE score as a function of UGT, UGPL, and SSAL that achieves synergy between UGT and UGPL, over the adaptation rounds (R1 and R2). The ECE score is computed (on the testing set) after each of the adaptation rounds.
    \textbf{Right.} The selected pseudo-labels corresponding to more certain detections have both low ECE score (computed before the round starts) and localization instability over the adaptation phase. The localization instability $\hat{l}_{n,m}$ for a selected pseudo-label is computed as: $\hat{l}_{n,m} = 1 -
     \left(\frac{1}{|\mathcal{T}_{n,m}|} \sum_{e} \mathcal{T}_{n,m}^{e} \right)$.
    }
\label{fig:analysisECE}
\end{figure}

\noindent \textbf{Uncertainty-guided pseudo-labelling for self-training.} As discussed above, the calibration can be considered as a measure of network's overall prediction uncertainty.
\textcolor{black}{Here, we attempt to discover the relationship between calibration and individual detection uncertainties.}
\textcolor{black}{To this end, we plot expected calibration error (ECE) score \cite{guo2017calibration} and output detection uncertainties (Fig. \ref{fig:analysisECE}).}

When we select pseudo-labels with more certain detections, the calibration error goes down significantly for this selected set. 
With this observation, we propose to select the pseudo-label $\tilde{\mathbf{y}}_{j,k}$ corresponding to detection $\widehat{\mathbf{y}}_{j,k}$ by utilizing uncertainty and detection consistency across $N$ inferences:
\begin{equation} 
    g_{j,k} = \mathbbm{1}[\hat{p}_{j,k} \ge \kappa_{1}] \mathbbm{1}[|\mathcal{T}_{j,k}| \ge \kappa_{2}],
    \label{eq:plSelec}
\end{equation}

where $\kappa_{1}$ and $\kappa_{2}$ are uncertainty and detection consistency thresholds.
\textcolor{black}{Some example detections considered for the pseudo-lables are shown in Fig.~\ref{fig:illustration_certain_vs_uncertain}.}
Once pseudo-labels are selected (Eq. \ref{eq:plSelec}), we use them to perform self-training as:
\begingroup
\normalsize
\begin{flalign}
\begin{split}
    \mathcal{L}_{pl}(\tilde{c}_{u,v}, \tilde{\mathbf{b}}_{u,v}) = \frac{1}{N_{pos}} \sum_{u,v} \mathbbm{1}_{\tilde{c}_{u,v} > 0}  \mathcal{L}_{cls}(\tilde{\mathbf{c}}_{u,v},c_{u,v}) \\
    + \frac{1}{N_{pos}} \sum_{u,v} \mathbbm{1}_{\tilde{c}_{u,v} > 0} \mathcal{L}
    _{box}(\tilde{\mathbf{b}}_{u,v}, \mathbf{b}_{u,v})
    \label{eq:UGST}
\end{split}
\end{flalign}
\endgroup

where $\tilde{c}_{u,v}, \tilde{\mathbf{b}}_{u,v}$ represents the class label and bounding box coordinates of the (selected) pseudo-label.
Compared to Eq.~(\ref{eq:det}),
in Eq.~(\ref{eq:UGST}), we back-propagate classification loss only for (selected) pseudo-label locations.

\noindent \textbf{Uncertainty-guided tiling for adversarial alignment.} Existing image and instance-level adversarial feature alignment suffer from interfering background and noisy object localization. We propose uncertainty-guided tiling for adversarial alignment; it mines relatively uncertain detected regions, as possible object-like regions, for the tiling process. Tiling anchored by uncertain object regions allows adversarial alignment to focus on potential, however, uncertain object-like region with context (see Fig.~\ref{fig:illustration_certain_vs_uncertain}). Specifically, if the averaged confidence $\hat{p}_{j,k}$ and the detection consistency $|\mathcal{T}_{j,k}|$, for a detection $\widehat{\mathbf{y}}_{j,k}$, is less than $\kappa_{1}$ and $\kappa_{2}$, respectively, it is mined as uncertain detection:
\begin{equation} 
    h_{j,k}= \mathbbm{1}[\hat{\kappa_{1}} \le \hat{p}_{j,k} < \kappa_{1}] \mathbbm{1}[|\mathcal{T}_{j,k}| < \kappa_{2}],
    \label{eq:tileselec}
\end{equation}
\begin{SCtable}[50][b]
\scriptsize
\centering
\renewcommand{\arraystretch}{1.5}
\tabcolsep=2.2pt\relax
\begin{tabular}{c|c|c}
\hline
   & SSAL\cite{munir2021ssal}  & SSAL\textsuperscript{\textdagger}   \\ \hline \hline
   R-1 & 86.2 & \textbf{90.2}   \\ \hline
 R-2 &   82.1 & \textbf{87.5} \\ \hline
\end{tabular}
\captionsetup{justification=justified}
\caption{\small Comparison of pseudo-labelling accuracy across two adaptation rounds R-1 \& R-2. SSAL (Eq.~\ref{eq:plSelec}) and SSAL\textsuperscript{\textdagger} (Eq.~\ref{eq:plSelec_3}) on \textbf{Sim10k $\rightarrow$ Cityscapes}.}
\label{tab:placcur}
\end{SCtable}
where $\hat{\kappa_{1}}$ is the lower limit to filter detections with very low uncertainty altogether since they potentially contain background clutter. 
Particularly, given $\acute{\mathbf{b}}_{j,k}$ as bounding box for detection $\acute{\mathbf{y}}_{j,k}$ for which $h_{j,k} = 1$, we crop a tile (region) $T_{i}$ of scale $W$ times as that of the detected bounding box. For source image, we randomly extract a tile $S_{i}$ around the ground-truth bounding box. 
\textcolor{black}{For $S_{i}$, we crop the source image of random sizes at random locations. Those cropped regions are declared tiles that contain at least 60\% of the image and at least one ground truth object. 
The term "around" for $S_{i}$ corresponds to the presence of at least one ground truth object for selection of tiles.}
After resizing both $T_{i}$ and $S_{i}$ to input image size, we perform the adversarial alignment as:

\begingroup
\small
\begin{flalign}
\begin{split}
    \mathcal{L}_{advT}(S_i, T_i) = -\sum_{u,v} q \log(D_{advT}(F_{T}^s)_{u,v}) \\
    + (1-q) \log(1-D_{advT}(F_{T}^t)_{u,v}),
    \label{eq:ugt}
\end{split}
\end{flalign}
\endgroup

where $F_{T}^s$ and $F_{T}^t$ are the feature maps for $S_{i}$ and $T_{i}$, respectively.

\begin{figure}[t]
    \centering
    \includegraphics[width=1.0\linewidth]{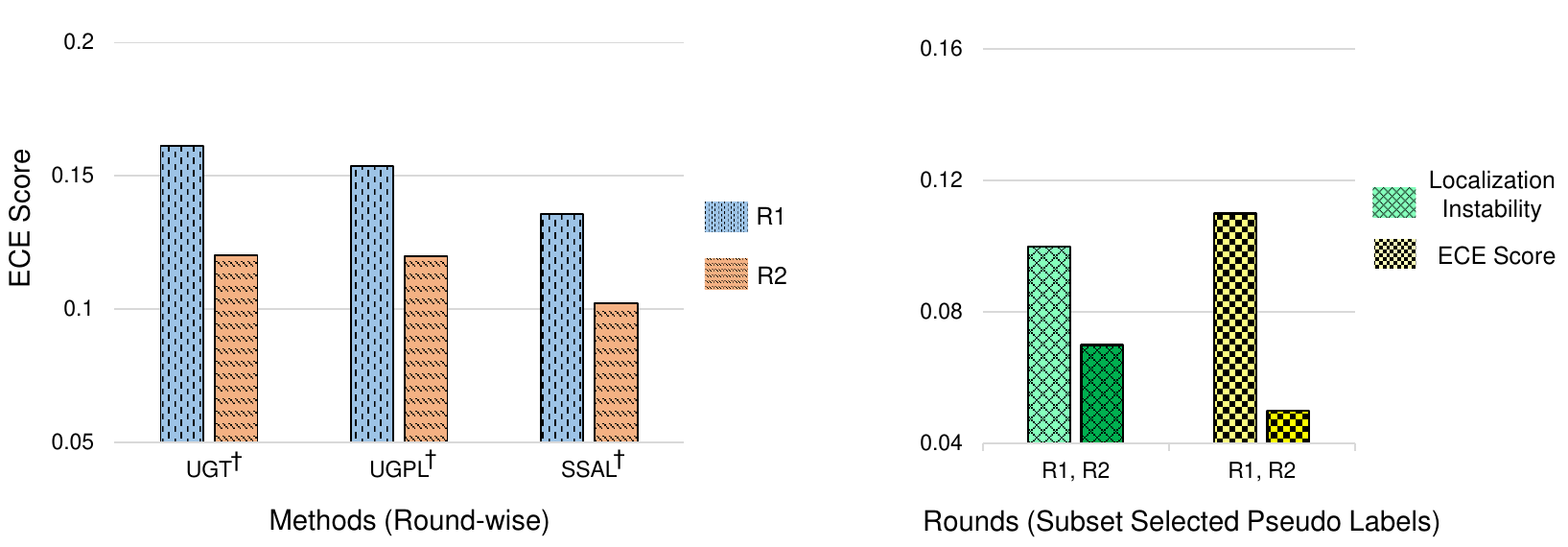}
    %\vspace{-0.3cm}
    \captionsetup{justification=justified}
    \caption{\small  
    \textbf{Left.} Analyzing ECE score as a function of UGT\textsuperscript{\textdagger}, UGPL\textsuperscript{\textdagger}, and SSAL\textsuperscript{\textdagger} through the adaptation rounds.
    \textbf{Right.} The selected pseudo-labels corresponding to more certain detections have both low ECE score (computed before the round starts) and localization instability through the adaptation phases. Note that, compared to SSAL, SSAL\textsuperscript{\textdagger} further improves model's calibration and localization stability.
    }
\label{fig:analysisECEssaldagger}
\end{figure}
%%add

\noindent \textbf{Discussion.} 
We analyze the impact on model's calibration through the adaptation phase after (1) selecting pseudo-labels with more certain detections (UGPL), (2) performing tiling on relatively uncertain detections (UGT), and (3) achieving the the synergy between UGPL and UGT (our method).
Model's calibration can be measured with Expected Calibration Error (ECE) score. 
We compute ECE score by considering both the confidence and the regression branch of the detector \cite{kueppers_2020_CVPR_Workshops}~\footnote{\small{Description on how ECE score is computed for detector is included in supplementary material.}}.  
Fig.~ \ref{fig:analysisECE} reveals that UGPL results in decreasing ECE score, and similarly (UGT) allows reducing the same even further. Finally, the synergy between UGPL and UGT achieves the lowest ECE score, significantly alleviating the impact of poor model's calibration under domain shift.

\noindent \textbf{Training objective.} We combine Eq.(\ref{eq:det}), Eq.(\ref{eq:UGST}), and Eq.(\ref{eq:ugt}) into a joint loss as $\mathcal{L} =  \mathcal{L}_{det} +  \mathcal{L}_{pl} +  \mathcal{L}_{adv}$ and optimize it to adapt the source model to the target domain. See supplementary material for the training pseudo-code of SSAL.

\subsection{SSAL\textsuperscript{\textdagger}}
\noindent \textbf{Revisiting uncertainty quantification.}
We note that SSAL \cite{munir2021ssal} relies on averaged class confidences as a surrogate measure of detection model uncertainty in its class assignment and object localization. 
It doesn't take into account the spread of the distribution, and so could be misleading for predictions with relatively greater localization uncertainty.
To this end, we revisit the uncertainty quantification (in sec.\ref{sec:SSAL}) and introduce variance across class confidences with the averaged class confidences. Specifically, given $\mathcal{T}_{n,m}$ (Eq. \ref{eq:tset}), $\widehat{p}_{n,m}^{e}$ (class prediction confidence of $e_{th}$ detection in $\mathcal{T}_{n,m}$) and $\hat{p}_{n,m}$ (Eq. \ref{eq:uncer}) we estimate variance across class prediction confidence for $m_{th}$ detection in $n_{th}$ inference as:
\begin{equation}
    \label{eq:uncersd}
     \hat{s}^2_{n,m} = \frac{\sum (\widehat{p}_{n,m}^{e}-\hat{p}_{n,m})^2}{|\mathcal{T}_{n,m}|}.
\end{equation}

\textcolor{black}{Since the variance across class confidences (Eq. \ref{eq:uncersd}) is much better estimate of the predictive uncertainty, using them in conjunction with  averaged class confidences (Eq. \ref{eq:uncer}) will allow us to further improve the synergy between self-training and adversarial alignment through facilitating more accurate pseudo-labelling and informed tiling (sec.\ref{sec:SSAL}).}

\noindent \textbf{Uncertainty-guided pseudo-labelling with new constraint.}
SSAL \cite{munir2021ssal} selects pseudo-labels using averaged class confidences as uncertainty measure and detection consistency (Eq. (\ref{eq:plSelec})) for self-training. This allows us to choose accurate pseudo-labels over sole confidence-based criterion, which is crucial for the effective adaptation and also improves model calibration under domain shift.
To further improve the selection of pseudo-labels, we propose to use average class confidences and variance across class confidences as model's detection uncertainty along with the detection consistency: 
\begin{equation} 
    g_{j,k} = \mathbbm{1}[\hat{s}^2_{j,k} \le \kappa_{0}] \mathbbm{1}[\hat{p}_{j,k} \ge \kappa_{1}] \mathbbm{1}[|\mathcal{T}_{j,k}| \ge \kappa_{2}],
    \label{eq:plSelec_3}
\end{equation}

where $\kappa_{0}$ is a threshold on variance constraint. Eq.(\ref{eq:plSelec_3}) allows us to select potentially more accurate pseudo-labels (see Tab. \ref{tab:placcur}), fulfilling the criteria of high average confidence in-tandem with low variance. The higher accuracy of pseudo-labels facilitates improved adaptation performance across various challenging scenarios (Tabs.~\ref{tab:cstofog} \& \ref{tab:simkittitocs}). Also, we show in Tab.~\ref{tab:abmodules} that, pseudo-label selection according to Eq.(\ref{eq:plSelec_3}) demonstrates better localization over different ranges of IoU.

% \noindent \textbf{(b) For Tiling.}
\noindent \textbf{Uncertainty-guided tiling with extended set.}
In SSAL\cite{munir2021ssal}, the detected regions satisfying the criteria in Eq. \ref{eq:tileselec}, are used to extract tiles for adversarial learning. We observe that the regions that fail the uncertainty constraint but satisfy the detection consistency ($\mathcal{|T|}$) constraint were not utilized for extracting tiles. This rather limits the space of uncertain detections (possibly containing some object information) that can be potentially exploited for enhanced adversarial alignment. Formally, we choose a region for extracting a tile that fulfills the following criteria:
\begingroup
\small
\begin{flalign}
\begin{split}
    h_{j,k} =  \mathbbm{1}[\hat{\kappa_{1}} \le \hat{p}_{j,k} < \kappa_{1}], 
    \label{eq:tile_1}
\end{split}
\end{flalign}
\endgroup

Further, along with the extracted tiles using  Eq.(\ref{eq:tile_1}), we also randomly sample full image in the mini-batch to extend scale information. We note that, in Tab.~\ref{tab:abmodules}, uncertainty-guided tiling based on Eq.(\ref{eq:tile_1}) and randomly sampled full image achieves better performance than the same relying on Eq.(\ref{eq:tileselec}) (described in sec.~\ref{sec:SSAL}). 

\noindent \textbf{Discussion.} Fig.~\ref{fig:analysisECEssaldagger} shows the impact on model's calibration via the adaptation phase after selecting the pseudo-labels with Eq.(\ref{eq:plSelec_3}) (UGPL\textsuperscript{\textdagger}), performing tiling on uncertain regions with Eq.(\ref{eq:tile_1}) (UGT\textsuperscript{\textdagger}), and achieving the synergy between UGPL\textsuperscript{\textdagger} and UGT\textsuperscript{\textdagger}  (SSAL\textsuperscript{\textdagger}). We observe that both UGPL\textsuperscript{\textdagger} and UGT\textsuperscript{\textdagger} further decrease the ECE score over their counterpart versions UGPL and UGT. Likewise, the synergy between UGPL\textsuperscript{\textdagger} and UGT\textsuperscript{\textdagger}, termed as SSAL\textsuperscript{\textdagger}, results in the lowest ECE score compared to SSAL \cite{munir2021ssal}.

% \begin{SCtable*}[50][t]
\begin{table*}[t]
\scriptsize
\centering
\renewcommand{\arraystretch}{1.3}
\tabcolsep=2.2pt\relax

\begin{tabular}{c|c|c|c|c|c|c|c|c|c|l}
\hline
\textbf{Method}           & \textbf{person} & \textbf{rider} & \textbf{car}  & \textbf{truck} & \textbf{bus}  & \textbf{train} & \textbf{mbike} & \textbf{bicycle} & \textbf{mAP@0.5} & \textbf{SO / Gain} \\ \hline \hline
\multicolumn{11}{c}{\textbf{Two-stage object detector}}                                                                                   \\ \hline \hline
DAF    \cite{chen2018domain}                  & 25.0            & 31.0           & 40.5          & 22.1           & 35.3          & 20.2           & 20.0           & 27.1             & 27.6             & 18.8 / 8.8           \\\cline{1-1} \cline{2-11} 
SW-DA  \cite{saito2019strong}   & 29.9            & 42.3           & 43.5          & 24.5           & 36.2          & 32.6           & 30.0           & 35.3             & 34.3             & 20.3 / 14.0          \\\cline{1-1} \cline{2-11} 

DAM            \cite{he2019multi}   & 30.8            & 40.5           & 44.3          & 27.2           & 38.4          & 34.5           & 28.4           & 32.2             & 34.6             & 18.8 / 16.7     

\\\cline{1-1} \cline{2-11} 

CR-DA   \cite{xu2020exploring}  & 32.9            & 43.8           & 49.2          & 27.2           & 45.1          & 36.4           & 30.3           & 34.6 & 37.4 & 22.0 / 15.4       \\\cline{1-1} \cline{2-11} 
CF-DA     \cite{zheng2020cross}   & \text{34.0}   & 46.9           & 52.1          & \text{30.8}  & 43.2            & \text{29.9}  & 34.7           & 37.4             & 38.6             & 20.8 / 17.8          \\\cline{1-1} \cline{2-11}  

% \rowcolor[HTML]{33e9ff}
ATF      \cite{he2020domain}    & 34.6            & 47.0           & 50.0          & 23.7           & \text{43.3}          & 38.7           & 33.4           & 38.8             & 38.7             & 20.3 / 18.4          \\\cline{1-1} \cline{2-11} 

HTCN      \cite{chen2020harmonizing}    & 33.2            & 47.5           & 47.9          & 31.6           & \text{47.4}          & 40.9           & 32.3           & 37.1             & 39.8             & 20.3 / 19.5          \\\cline{1-1} \cline{2-11} 
UADA     \cite{nguyen2020uada}                             & 34.2            & \text{48.9}  & \text{52.4} & 30.3           & 42.7          & 46.0             & \text{33.2}  & \text{36.2}    & 40.5             & 20.3 / 20.2          \\\cline{1-1} \cline{2-11}  
SAPNet      \cite{li2020spatial}        & 40.8            & 46.7           & \text{59.8} & 24.3           & 46.8          & 37.5           & 30.4           & \text{40.7}    &  \text{40.9}             & 20.3 / \textbf{20.6} 
\\\cline{1-1} \cline{2-11}  
% \rowcolor[HTML]{33e9ff}
D-adapt      \cite{jiang2021decoupled}        & 44.9            & 54.2           & \text{61.7} & 25.6           & 36.3          & 24.7           & 37.3           & \text{46.1}    &  \text{41.3}             & 23.4 / \text{17.9}        
\\ \hline \hline
\multicolumn{11}{c}{\textbf{One-stage object detector}} 
\\ \hline \hline

\multicolumn{1}{c|}{Source Only}                & 31.7            & 31.7           & 34.6          & 5.9            & 20.3          & 2.5            & 10.6           & 25.8             & 20.4             & -         \\\cline{1-1} \cline{2-11} 
Baseline         \cite{hsu2020every}              & 38.7            & 36.1           & 53.1          & 21.9           & 35.4          & 25.7           & 20.6           & 33.9             & 33.2             & 18.4 / 14.8           \\\cline{1-1} \cline{2-11} 

EPM         \cite{hsu2020every}       & 41.9            & 38.7           & 56.7          & 22.6           & 41.5          & \text{26.8}           & 24.6           & 35.5             & 36.0             & 18.4 / 17.6   \\\cline{1-1} \cline{2-11}  
\multicolumn{1}{c|}{\text{Ours (SSAL) \cite{munir2021ssal}}} &           \text{45.1}   & \text{47.4}  & \text{59.4} & \text{24.5}  & \text{50.0} & 25.7           & \text{26.0}    & \text{38.7}    & \text{39.6}    & 20.4 / \text{19.2}  
\\\cline{1-1} \cline{2-11}   
\multicolumn{1}{c|}{Ours (\textbf{SSAL}\textsuperscript{\textdagger})}                             & \text{46.3}            & 45.8           & \text{59.4}          & \text{24.8}           & 45.3          & \text{30.6}           & \text{26.7}           & \text{39.7}             & \text{39.8}             & 20.4 / \textbf{19.4}           
\\\cline{1-1} \cline{2-11}   
\multicolumn{1}{c|}{Oracle}                             & 47.4            & 40.8           & 66.8          & 27.2           & 48.2          & 32.4           & 31.2           & 38.3             & 41.5             & -             

\\ \hline
\end{tabular}
%%\vspace{0.3cm}
\captionsetup{justification=justified}
\caption{\small \textbf{Cityscapes $\rightarrow$ Foggy Cityscapes}:  
SSAL\textsuperscript{\textdagger} achieves an absolute gain of 19.4\% over the source only model and outperforms most recent one-stage domain adaptive detector (EPM). SSAL\textsuperscript{\textdagger} further improves over SSAL by an absolute margin of 0.2\% both in mAP and gain. $SO$ refers to source only. 
The best results are bold-faced.}
\label{tab:cstofog}
\end{table*}

\begin{table}[!ht]
\scriptsize
\centering
\renewcommand{\arraystretch}{1.2}
\tabcolsep=1.9pt\relax
\begin{tabular}{c|c|c|c|c}

\multicolumn{1}{c}{\textbf{}}                                           & \multicolumn{2}{c|}{\textbf{Sim10k} $\rightarrow$ \textbf{CS}}                                                                                         & \multicolumn{2}{c}{\textbf{KITTI} $\rightarrow$ \textbf{CS}}                                                    \\ \hline

\textbf{Method}                                     & \begin{tabular}[c]{@{}c@{}}  \textbf{AP @ 0.5}\end{tabular} & \textbf{SO / Gain}   & \multicolumn{1}{c|}{\begin{tabular}[c]{@{}c@{}}\ \textbf{AP @ 0.5}\end{tabular}} & \multicolumn{1}{c}{\textbf{SO / Gain}} \\ \hline
\multicolumn{5}{c}{\textbf{Two   Stage Object Detector}}                                                                                                           \\ \hline
DAF  \cite{chen2018domain}         & 39.0                                                                                & 30.1 / 8.9                            & 38.5                                                                                                     & 30.2 / 8.3                                               \\ \hline
SC-DA   \cite{zhu2019adapting}     & 43.0                                                                                & 34.0 / 9.0                            & \textbf{42.5}                                                                                                     & 37.4 / 5.1                                               \\ \hline
MAF    \cite{he2019multi}          & 41.1                                                                                & 30.1 / \textbf{11.0} & 41.0                                                                                                     & 30.2 / \textbf{10.8}                                              \\ \hline
CF-DA     \cite{zheng2020cross}    & 43.8                                                                                & 35.0 / 8.8                            & -                                                                                                        & -                                                        \\ \hline
% \rowcolor[HTML]{33e9ff}
ATF    \cite{he2020domain} & 42.8                                                                                & 34.6 / 8.2                           & -                                                                                                        & -                                                        \\ \hline
HTCN    \cite{chen2020harmonizing} & 42.5                                                                                & 34.6 / 7.9                            & -                                                                                                        & -                                                        \\ \hline
SAPNet  \cite{li2020spatial}       & \text{44.9}                                                      & 34.6 / 10.3                           & -                                                                                                        & -                                                        \\ \hline
UADA     \cite{nguyen2020uada}     & 42.0                                                                                & 34.6 / 7.4                            & -                                                                                                        & -                                                        \\ \hline
% \rowcolor[HTML]{33e9ff}
D-adapt     \cite{jiang2021decoupled}     & \textbf{50.3}                                                                                & 34.6 / 15.7                            & -                                                                                                        & -                                                        \\ \hline
\multicolumn{5}{c}{\textbf{One Stage Object Detector}}                                                                                                                                                                                                                                                                                                \\ \hline
Source Only                                         & 38.0                                                                                & -                                     & 34.9                                                                                                     & -                                                         \\ \hline
Baseline \cite{hsu2020every}                                            & 46.0                                                                                & 39.8 / 6.2                            & 39.1                                                                                                     & 34.4 / 4.7                                               \\ \hline
EPM  \cite{hsu2020every}           & 49.0                                                                                & 39.8 / 9.2                            & 43.2                                                                                                     & 34.4 / 8.8                                               \\ \hline
Ours (SSAL) \cite{munir2021ssal}                                               & \text{51.8}                                                      & 38.0 / \text{13.8} & \text{45.6}                                                                                            & 34.9 / \text{10.7}                                     \\ \hline

Ours (\textbf{SSAL}\textsuperscript{\textdagger})                                               & \textbf{53.0}                                                      & 38.0 / \textbf{15.0} & \textbf{46.7}                                                                                            & 34.9 / \textbf{11.8}                                     \\ \hline

Oracle                                              & 69.7                                                                                & -                                     & 69.7                                                                                                     & -                                                        \\ \hline
\end{tabular}
% %\vspace{0.3cm}
\captionsetup{justification=justified}
\caption{
\small 
\textbf{Sim10k $\rightarrow$ Cityscapes}: SSAL\textsuperscript{\textdagger} outperforms one-stage and two-stage object detectors both in-terms of mAP(\%) and gain obtained over source.  
\textbf{KITTI $\rightarrow$ Cityscapes}:
SSAL\textsuperscript{\textdagger} is better than both the EPM and the existing SOTA methods with considerable margin in mAP. For qualitative figures (KITTI $\rightarrow$ Cityscapes), see supplementary material. $SO$ refers to source only.
The best results are bold-faced.
}
% \vspace{-0.3cm}
\label{tab:simkittitocs}
\end{table}

\begin{table*}
\scriptsize
\centering
\renewcommand{\arraystretch}{1.3}
\tabcolsep=2.2pt\relax
\begin{tabular}{ccccccccccc}
\hline
\multicolumn{1}{c|}{\textbf{Method}} & \multicolumn{1}{c|}{\textbf{person}} & \multicolumn{1}{c|}{\textbf{rider}} & \multicolumn{1}{c|}{\textbf{car}}  & \multicolumn{1}{c|}{\textbf{truck}} & \multicolumn{1}{c|}{\textbf{bus}}  & \multicolumn{1}{c|}{\textbf{train}} & \multicolumn{1}{c|}{\textbf{mcycle}} & \multicolumn{1}{c|}{\textbf{bicycle}} & \multicolumn{1}{c|}{\textbf{mAP@0.5}} & \textbf{SO / Gain} \\ \hline\hline
\multicolumn{11}{c}{\textbf{Two-stage object detector}}                                                                                   \\ \hline\hline
\multicolumn{1}{c|}{CR-DA-Faster \cite{xu2020exploring}}       & \multicolumn{1}{c|}{29.3}            & \multicolumn{1}{c|}{28.4}           & \multicolumn{1}{c|}{45.3}          & \multicolumn{1}{c|}{17.5}           & \multicolumn{1}{c|}{17.1}          & \multicolumn{1}{c|}{0}              & \multicolumn{1}{c|}{16.8}            & \multicolumn{1}{c|}{22.7}             & \multicolumn{1}{c|}{25.3}             & 23.4 / 1.9         \\ \hline
\multicolumn{1}{c|}{CR-SW-Faster \cite{xu2020exploring}}       & \multicolumn{1}{c|}{\textbf{31.4}}            & \multicolumn{1}{c|}{\textbf{31.3}}  & \multicolumn{1}{c|}{\textbf{46.3}}          & \multicolumn{1}{c|}{\textbf{19.5}}           & \multicolumn{1}{c|}{\textbf{18.9}}          & \multicolumn{1}{c|}{0}              & \multicolumn{1}{c|}{\textbf{17.3}}   & \multicolumn{1}{c|}{\textbf{23.8}}             & \multicolumn{1}{c|}{\textbf{26.9}}             & 23.4 / \textbf{3.5}         \\ \hline\hline
\multicolumn{11}{c}{\textbf{One-stage object detector}}                                                                                        \\ \hline\hline

\multicolumn{1}{c|}{Source Only}        & \multicolumn{1}{c|}{35.7}            & \multicolumn{1}{c|}{18.9}           & \multicolumn{1}{c|}{56.3}          & \multicolumn{1}{c|}{11.6}           & \multicolumn{1}{c|}{13.8}          & \multicolumn{1}{c|}{0}              & \multicolumn{1}{c|}{4.9}             & \multicolumn{1}{c|}{14.6}             & \multicolumn{1}{c|}{19.5}               & -         \\ \hline

\multicolumn{1}{c|}{Baseline \cite{hsu2020every}}        & \multicolumn{1}{c|}{35.6}            & \multicolumn{1}{c|}{21.5}           & \multicolumn{1}{c|}{56.6}          & \multicolumn{1}{c|}{13.1}           & \multicolumn{1}{c|}{13.7}          & \multicolumn{1}{c|}{0}              & \multicolumn{1}{c|}{9.6}             & \multicolumn{1}{c|}{18.1}             & \multicolumn{1}{c|}{21.0}               & 19.5 / 1.5         \\ \hline
% \rowcolor[HTML]{33e9ff}
\multicolumn{1}{c|}{Ours(\textbf{SSAL})}    & \multicolumn{1}{c|}{\text{44.2}}   & \multicolumn{1}{c|}{\text{25.5}}           & \multicolumn{1}{c|}{\text{62.7}} & \multicolumn{1}{c|}{\text{17.2}}  & \multicolumn{1}{c|}{\text{18.1}} & \multicolumn{1}{c|}{0}              & \multicolumn{1}{c|}{\text{12.4}}            & \multicolumn{1}{c|}{\text{24.8}}    & \multicolumn{1}{c|}{\text{25.6}}    & 19.5 / \text{6.1}         \\ \hline

\multicolumn{1}{c|}{Ours(\textbf{SSAL}\textsuperscript{\textdagger})}    & \multicolumn{1}{c|}{\textbf{46.5}}   & \multicolumn{1}{c|}{\textbf{26.9}}           & \multicolumn{1}{c|}{\textbf{64.4}} & \multicolumn{1}{c|}{\textbf{19.7}}  & \multicolumn{1}{c|}{\textbf{23.4}} & \multicolumn{1}{c|}{0}              & \multicolumn{1}{c|}{\textbf{13.8}}            & \multicolumn{1}{c|}{\textbf{25.6}}    & \multicolumn{1}{c|}{\textbf{27.5}}    & 19.5 / \textbf{8.0}         \\ \hline

\multicolumn{1}{c|}{Oracle}        & \multicolumn{1}{c|}{62.5}            & \multicolumn{1}{c|}{37.0}           & \multicolumn{1}{c|}{79.4}          & \multicolumn{1}{c|}{54.6}           & \multicolumn{1}{c|}{49.2}          & \multicolumn{1}{c|}{0}              & \multicolumn{1}{c|}{34.1}             & \multicolumn{1}{c|}{37.8}             & \multicolumn{1}{c|}{44.3}               & -         \\ \hline

\end{tabular}
%%\vspace{0.3cm}
\captionsetup{justification=justified}
\caption{\small \textbf{Cityscapes $\rightarrow$ BDD100k}:
Our method (SSAL\textsuperscript{\textdagger}) achieves an absolute gain of 8.0\% over the source only model and outperforms the baseline from one-stage domain adaptive detector (EPM) and two-stage domain adaptive methods. $SO$ refers to source only.
The best results are bold-faced. For qualitative figures (Cityscapes $\rightarrow$ BDD100k), see supplementary material.}
\label{tab:bdd}
\end{table*}

\begin{table*}[t]
\scriptsize
\centering
\renewcommand{\arraystretch}{1.1}
\tabcolsep=1.5pt\relax
\begin{tabular}{ccccccccccccccccccccccc}
\hline
\multicolumn{1}{c|}{\textbf{Method}} & \multicolumn{1}{c|}{\textbf{aero}} & \multicolumn{1}{c|}{\textbf{bicycle}} & \multicolumn{1}{c|}{\textbf{bird}} & \multicolumn{1}{c|}{\textbf{boat}} & \multicolumn{1}{c|}{\textbf{bottle}} & \multicolumn{1}{c|}{\textbf{bus}}  & \multicolumn{1}{c|}{\textbf{car}}  & \multicolumn{1}{c|}{\textbf{cat}}  & \multicolumn{1}{c|}{\textbf{chair}} & \multicolumn{1}{c|}{\textbf{cow}}  & \multicolumn{1}{c|}{\textbf{table}} & \multicolumn{1}{c|}{\textbf{dog}}  & \multicolumn{1}{c|}{\textbf{horse}} & \multicolumn{1}{c|}{\textbf{mbike}} & \multicolumn{1}{c|}{\textbf{person}} & \multicolumn{1}{c|}{\textbf{plant}} & \multicolumn{1}{c|}{\textbf{sheep}} & \multicolumn{1}{c|}{\textbf{sofa}} & \multicolumn{1}{c|}{\textbf{train}} & \multicolumn{1}{c|}{\textbf{tv}} & \multicolumn{1}{c|}{\textbf{mAP@0.5}} & \textbf{SO / Gain} \\ \hline
\multicolumn{23}{c}{\textbf{Two-stage object detector}}                                                           \\ \hline
\multicolumn{1}{c|}{SW-DA \cite{saito2019strong}}            & \multicolumn{1}{c|}{26.2}          & \multicolumn{1}{c|}{48.5}            & \multicolumn{1}{c|}{32.6}          & \multicolumn{1}{c|}{33.7}          & \multicolumn{1}{c|}{38.5}            & \multicolumn{1}{c|}{54.3}          & \multicolumn{1}{c|}{37.1}          & \multicolumn{1}{c|}{\textbf{18.6}} & \multicolumn{1}{c|}{34.8}           & \multicolumn{1}{c|}{58.3}          & \multicolumn{1}{c|}{17.0}            & \multicolumn{1}{c|}{12.5}          & \multicolumn{1}{c|}{33.8}           & \multicolumn{1}{c|}{65.5}           & \multicolumn{1}{c|}{61.6}            & \multicolumn{1}{c|}{52.0}            & \multicolumn{1}{c|}{9.3}            & \multicolumn{1}{c|}{24.9}          & \multicolumn{1}{c|}{54.1}           & \multicolumn{1}{c|}{49.1}           & \multicolumn{1}{c|}{38.1}            & 27.8 / 10.3        \\ \hline
\multicolumn{1}{c|}{TriWay FRCNN \cite{he2020domain}}    & \multicolumn{1}{c|}{\textbf{41.9}} & \multicolumn{1}{c|}{67.0}            & \multicolumn{1}{c|}{27.4}          & \multicolumn{1}{c|}{\textbf{36.4}} & \multicolumn{1}{c|}{41.0}            & \multicolumn{1}{c|}{48.5}          & \multicolumn{1}{c|}{\textbf{42.0}} & \multicolumn{1}{c|}{13.1}          & \multicolumn{1}{c|}{39.2}           & \multicolumn{1}{c|}{\textbf{75.1}} & \multicolumn{1}{c|}{\textbf{33.4}}   & \multicolumn{1}{c|}{7.9}           & \multicolumn{1}{c|}{41.2}           & \multicolumn{1}{c|}{56.2}           & \multicolumn{1}{c|}{51.4}            & \multicolumn{1}{c|}{\textbf{50.6}}   & \multicolumn{1}{c|}{\textbf{42.0}}  & \multicolumn{1}{c|}{25}            & \multicolumn{1}{c|}{52.1}           & \multicolumn{1}{c|}{39.1}           & \multicolumn{1}{c|}{42.1}            & 27.8 / 14.3        \\ \hline
\multicolumn{1}{c|}{SAPNet \cite{li2020spatial}}             & \multicolumn{1}{c|}{27.4}          & \multicolumn{1}{c|}{\textbf{70.8}}   & \multicolumn{1}{c|}{32}            & \multicolumn{1}{c|}{27.9}          & \multicolumn{1}{c|}{42.4}            & \multicolumn{1}{c|}{\textbf{63.5}} & \multicolumn{1}{c|}{47.5}          & \multicolumn{1}{c|}{14.3}          & \multicolumn{1}{c|}{\textbf{48.2}}  & \multicolumn{1}{c|}{46.1}          & \multicolumn{1}{c|}{31.8}            & \multicolumn{1}{c|}{17.9}          & \multicolumn{1}{c|}{\textbf{43.8}}  & \multicolumn{1}{c|}{68.0}           & \multicolumn{1}{c|}{\textbf{68.1}}   & \multicolumn{1}{c|}{49.0}            & \multicolumn{1}{c|}{18.7}           & \multicolumn{1}{c|}{20.4}          & \multicolumn{1}{c|}{\textbf{55.8}}  & \multicolumn{1}{c|}{\textbf{51.3}}  & \multicolumn{1}{c|}{\textbf{42.2}}   & 27.8 / \textbf{14.4}        \\ \hline
\multicolumn{1}{c|}{II-DAOD\cite{wu2021instance}}         & \multicolumn{1}{c|}{41.5}          & \multicolumn{1}{c|}{52.7}            & \multicolumn{1}{c|}{\textbf{34.5}} & \multicolumn{1}{c|}{28.1}          & \multicolumn{1}{c|}{\textbf{43.7}}   & \multicolumn{1}{c|}{58.5}          & \multicolumn{1}{c|}{41.8}          & \multicolumn{1}{c|}{15.3}          & \multicolumn{1}{c|}{40.1}           & \multicolumn{1}{c|}{54.4}          & \multicolumn{1}{c|}{26.7}            & \multicolumn{1}{c|}{\textbf{28.5}} & \multicolumn{1}{c|}{37.7}           & \multicolumn{1}{c|}{\textbf{75.4}}  & \multicolumn{1}{c|}{63.7}            & \multicolumn{1}{c|}{48.7}            & \multicolumn{1}{c|}{16.5}           & \multicolumn{1}{c|}{\textbf{30.8}} & \multicolumn{1}{c|}{54.5}           & \multicolumn{1}{c|}{48.7}           & \multicolumn{1}{c|}{42.1}            & 27.8 / 14.3        \\ \hline
\multicolumn{23}{c}{\textbf{One-stage object detector}}                                                \\ \hline
\multicolumn{1}{c|}{Source Only}     & \multicolumn{1}{c|}{33.2}          & \multicolumn{1}{c|}{28.1}            & \multicolumn{1}{c|}{20.5}          & \multicolumn{1}{c|}{18.0}          & \multicolumn{1}{c|}{21.4}            & \multicolumn{1}{c|}{29.8}          & \multicolumn{1}{c|}{36.7}          & \multicolumn{1}{c|}{8.0}           & \multicolumn{1}{c|}{28.9}           & \multicolumn{1}{c|}{8.2}           & \multicolumn{1}{c|}{17.8}            & \multicolumn{1}{c|}{4.9}           & \multicolumn{1}{c|}{20.3}           & \multicolumn{1}{c|}{32.3}           & \multicolumn{1}{c|}{41.5}            & \multicolumn{1}{c|}{40.7}            & \multicolumn{1}{c|}{1.0}            & \multicolumn{1}{c|}{18.8}          & \multicolumn{1}{c|}{37.0}           & \multicolumn{1}{c|}{42.6}           & \multicolumn{1}{c|}{24.5}            &                    \\ \hline
\multicolumn{1}{c|}{Baseline\cite{hsu2020every}}        & \multicolumn{1}{c|}{24.4}          & \multicolumn{1}{c|}{39.1}            & \multicolumn{1}{c|}{18.6}          & \multicolumn{1}{c|}{16.9}          & \multicolumn{1}{c|}{27.4}            & \multicolumn{1}{c|}{48.1}          & \multicolumn{1}{c|}{34.6}          & \multicolumn{1}{c|}{0.6}           & \multicolumn{1}{c|}{32.0}           & \multicolumn{1}{c|}{36.0}          & \multicolumn{1}{c|}{15.8}            & \multicolumn{1}{c|}{7.6}           & \multicolumn{1}{c|}{16.2}           & \multicolumn{1}{c|}{56.8}           & \multicolumn{1}{c|}{47.7}            & \multicolumn{1}{c|}{36.7}            & \multicolumn{1}{c|}{\textbf{10.5}}  & \multicolumn{1}{c|}{22.5}          & \multicolumn{1}{c|}{38.2}           & \multicolumn{1}{c|}{36.9}           & \multicolumn{1}{c|}{28.3}            & 24.5 / 3.8         \\ \hline
\multicolumn{1}{c|}{WST-BSR\cite{kim2019self}}         & \multicolumn{1}{c|}{28.0}          & \multicolumn{1}{c|}{\textbf{64.5}}   & \multicolumn{1}{c|}{23.9}          & \multicolumn{1}{c|}{19.0}          & \multicolumn{1}{c|}{21.9}            & \multicolumn{1}{c|}{\textbf{64.3}} & \multicolumn{1}{c|}{\textbf{43.5}} & \multicolumn{1}{c|}{16.4}          & \multicolumn{1}{c|}{\textbf{42.2}}  & \multicolumn{1}{c|}{25.9}          & \multicolumn{1}{c|}{\textbf{30.5}}   & \multicolumn{1}{c|}{\textbf{7.9}}  & \multicolumn{1}{c|}{25.5}           & \multicolumn{1}{c|}{\textbf{67.6}}  & \multicolumn{1}{c|}{54.5}            & \multicolumn{1}{c|}{36.4}            & \multicolumn{1}{c|}{10.3}           & \multicolumn{1}{c|}{\textbf{31.2}} & \multicolumn{1}{c|}{\textbf{57.4}}  & \multicolumn{1}{c|}{43.5}           & \multicolumn{1}{c|}{35.7}            & 26.7 / 9.0         \\ \hline
% \rowcolor[HTML]{33e9ff}
\multicolumn{1}{c|}{Ours (\textbf{SSAL})}    & \multicolumn{1}{c|}{\text{38.9}} & \multicolumn{1}{c|}{37.9}            & \multicolumn{1}{c|}{\text{30.0}} & \multicolumn{1}{c|}{\text{26.1}} & \multicolumn{1}{c|}{\text{35.1}}   & \multicolumn{1}{c|}{42.0}          & \multicolumn{1}{c|}{35.6}          & \multicolumn{1}{c|}{\text{15.0}} & \multicolumn{1}{c|}{37.1}           & \multicolumn{1}{c|}{\textbf{50.8}} & \multicolumn{1}{c|}{24.5}            & \multicolumn{1}{c|}{6.2}           & \multicolumn{1}{c|}{\textbf{27.3}}  & \multicolumn{1}{c|}{51.6}           & \multicolumn{1}{c|}{\textbf{62.7}}   & \multicolumn{1}{c|}{\text{39.4}}   & \multicolumn{1}{c|}{10.1}            & \multicolumn{1}{c|}{22.9}          & \multicolumn{1}{c|}{47.3}           & \multicolumn{1}{c|}{\text{49.8}}  & \multicolumn{1}{c|}{\text{34.5}}   & 24.5 / \text{10.0}        \\ \hline
\multicolumn{1}{c|}{Ours (\textbf{SSAL}\textsuperscript{\textdagger})}    & \multicolumn{1}{c|}{\textbf{41.2}} & \multicolumn{1}{c|}{46.4}            & \multicolumn{1}{c|}{\textbf{30.8}} & \multicolumn{1}{c|}{\textbf{29.0}} & \multicolumn{1}{c|}{\textbf{36.7}}   & \multicolumn{1}{c|}{48.2}          & \multicolumn{1}{c|}{37.7}          & \multicolumn{1}{c|}{\textbf{17.4}} & \multicolumn{1}{c|}{37.3}           & \multicolumn{1}{c|}{\text{50.2}} & \multicolumn{1}{c|}{25.9}            & \multicolumn{1}{c|}{6.9}           & \multicolumn{1}{c|}{\text{26.6}}  & \multicolumn{1}{c|}{56.1}           & \multicolumn{1}{c|}{\text{62.6}}   & \multicolumn{1}{c|}{\textbf{42.8}}   & \multicolumn{1}{c|}{8.4}            & \multicolumn{1}{c|}{26.7}          & \multicolumn{1}{c|}{45.5}           & \multicolumn{1}{c|}{\textbf{54.3}}  & \multicolumn{1}{c|}{\textbf{36.5}}   & 24.5 / \textbf{12.0}        \\ \hline

\end{tabular}
\captionsetup{justification=justified}
\caption{\small \textbf{PASCALVOC $\rightarrow$ Clipart1k}:
SSAL\textsuperscript{\textdagger} achieves an absolute gain of 12.0\% over the source only model and outperforms the SSD based one-stage domain adaptive detector (WST-BSR). $SO$ refers to source only.
The best results are bold-faced.}
\label{tab:voctoclip}
\end{table*}

\section{Experiments}

\label{sec:exp}
%\vspace{-0.2cm}

\noindent \textbf{Datasets.} \textbf{Cityscapes} \cite{Cordts2016Cityscapes} dataset features images of road and street scenes and offers 2975 and 500 examples for training and validation, respectively. It comprises following categories: \textit{person, rider, car, truck, bus, train, motorbike, and bicycle}.
\textbf{Foggy Cityscapes} \cite{sakaridis2018semantic} dataset is constructed using Cityscapes dataset by simulating foggy weather utilizing depth maps provided in Cityscapes with three levels of foggy weather.
\textbf{Sim10k} \cite{johnson2017driving} dataset is a collection of synthesized images, comprising 10K images and their corresponding bounding box annotations. 
\textbf{KITTI} \cite{geiger2012we} dataset bears resemblance to Cityscapes as it features images of road scenes with wide view of area, except that KITTI images were captured with a different camera setup.
Following existing works, we consider car class for experiments when adapting from KITTI or Sim10k.
\textbf{PASCALVOC} \cite{everingham2010pascal} 
is a well-known dataset in object detection literature, containing 20 categories. This dataset offers real images with bounding box and category level information. 
Following the protocol in \cite{saito2019strong}, we use PASCAL VOC 2007 and 2012 training and validation sets as training data.
\textbf{Clipart1k} \cite{inoue2018cross} contains artistic images with 1k samples. This dataset has same 20 categories as in PASCAL VOC\cite{everingham2010pascal}. Following \cite{saito2019strong}, we utilize all images for training (without annotations) and testing.
\textbf{BDD100k} \cite{yu2018bdd100k} is a large-scale dataset, containing 100k images with bounding box and class level annotations. Out of these images, 70k are in training set and 10k is in the validation set. Following \cite{xu2020exploring}, we make a subset of 36.7k images from the training set and 5.2k images from the validation set that has daylight conditions and used with common categories as in Cityscapes.
 
\noindent \textbf{Implementation and Training Details.}
We train FCOS \cite{tian2019fcos}, fully convolutional one-stage object detector, over the source domain data.
During the adaptation process, using the source-trained model, we iterate over two steps: UGPL/UGPL\textsuperscript{\textdagger} and UGT/UGT\textsuperscript{\textdagger}. 
Following \cite{zou2019confidence, zou2018unsupervised} we define going over these two steps once as \textit{Domain Adaptation Round} or just \textit{Round}. 
In all experiments for uniformity, we use three rounds. 
Since initially pseudo-labelling accuracy is likely poor, following \cite{NEURIPS2019_categoryanch}, we perform adversarial domain adaptation (using UGT\textsuperscript{\textdagger}), in a round called R0. 
In the next two rounds, R1 and R2, we apply both the self-training and adversarial domain adaptation using UGPL/UGPL\textsuperscript{\textdagger} and UGT/UGT\textsuperscript{\textdagger}, respectively. 
For extracting tile around uncertain detection, a five times larger region is cropped around the center location. Height and width are re-adjusted to make the extracted tile square, so that during the resizing in any later stage the aspect ratio of any object in tile remains unaffected. 

We set mini-batch size to 3. 
The learning rate is set to $5\times 10^{-3}$ during training of source model and R0 round and then reduced to $1\times 10^{-3}$ during the R1 and R2. 
R1 and R2 consist of $10K$ iterations, R0 however performs 5K iterations. IoU threshold $\gamma$ is set to 0.5.
We use $N=10$ MC-dropout inferences, with dropout rate set to 10\%. All experiments are performed using a single GPU (Quadro RTX 6000).
 $\kappa_{1}$ and $\kappa_{2}$, uncertainty and detection consistency thresholds, are both set to 0.5, indicating object same class prediction and location should occur at-least 50\% of times. $\kappa_{0}$ and 
 $\hat{\kappa_{1}}$ is set to 0.1 to filter highly uncertain detections.
All training and testing images are resized such that their shorter side has 800 pixels. 
We use VGG-16\cite{simonyan2014very} as backbone for all adaptations except for PASCALVOC to Clipart1k for which ResNet101\cite{he2016deep} is used as backbone in the current literature \cite{saito2019strong,xu2020exploring}.

%\vspace{-0.1cm}
\subsection{Comparison with the state-of-the-art}
%\vspace{-0.1cm}
For all the domain adaptation experiments we compare both the existing state-of-the-art (SOTA), one-stage and two-stage object detectors using the same feature backbone.
Results are compared in terms of mAP(\%), class-wise APs(\%), and gain (\%) achieved over a source only model.
To better understand the effectiveness of our domain-adaptive algorithm, we also report results on \textbf{Baseline}, which is FCOS \cite{tian2019fcos} along with the global-level feature alignment. 

\begin{figure*}[t]
    \centering
    \includegraphics[width=0.8\linewidth]{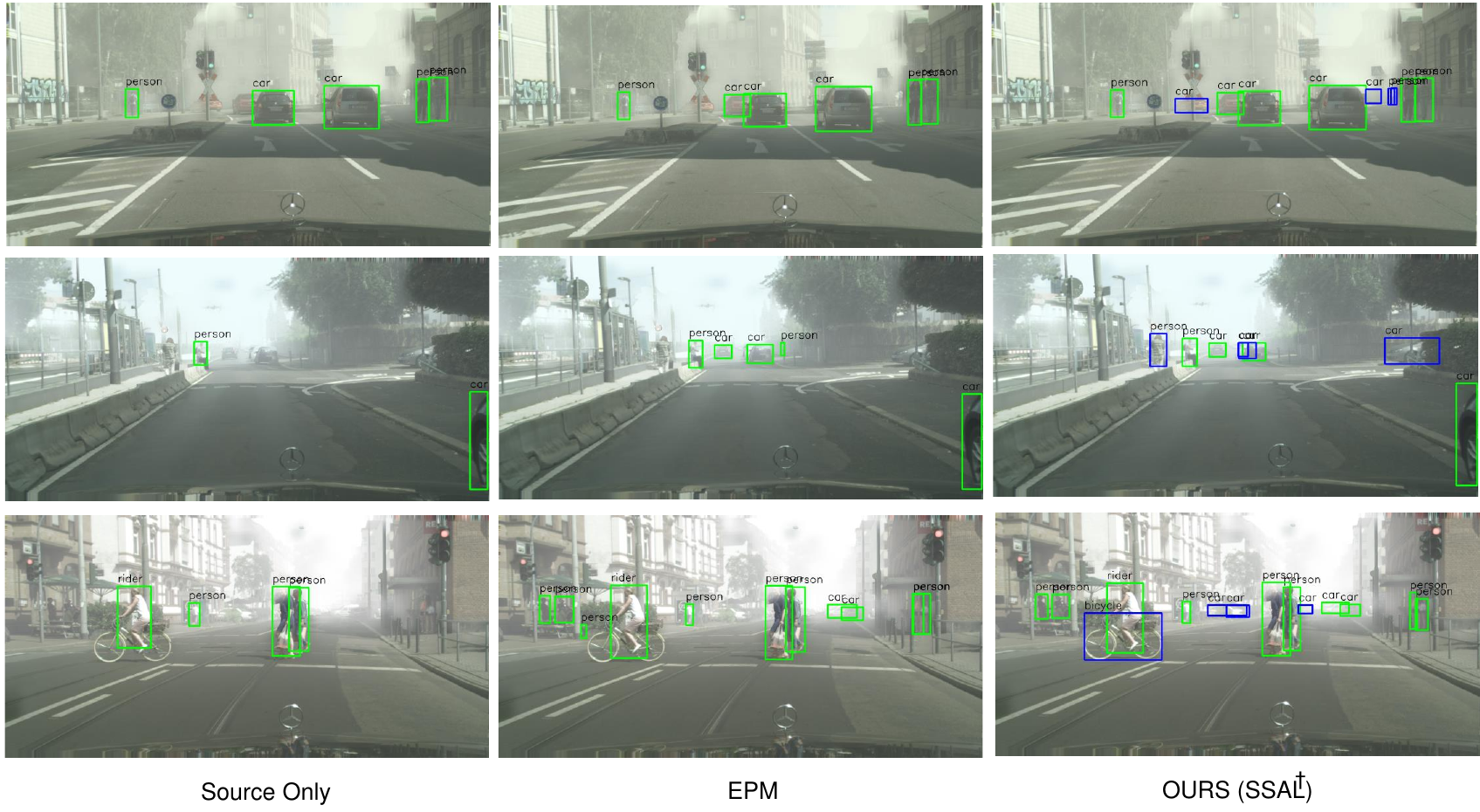}
    \captionsetup{justification=justified}
    \caption{\small \textbf{Cityscapes $\rightarrow$ Foggy Cityscapes:} Detections missed by the EPM and found by SSAL\textsuperscript{\textdagger} are shown in \textcolor{black}{Blue}. Compared to EPM~\cite{hsu2020every}, SSAL\textsuperscript{\textdagger} better localizes and detects objects, especially the far ones, under foggy conditions.}
    %\vspace{-0.3cm}
    \label{fig:qual}
\end{figure*}

\begin{figure*}[t]
    \centering
    \includegraphics[width=0.8\linewidth]{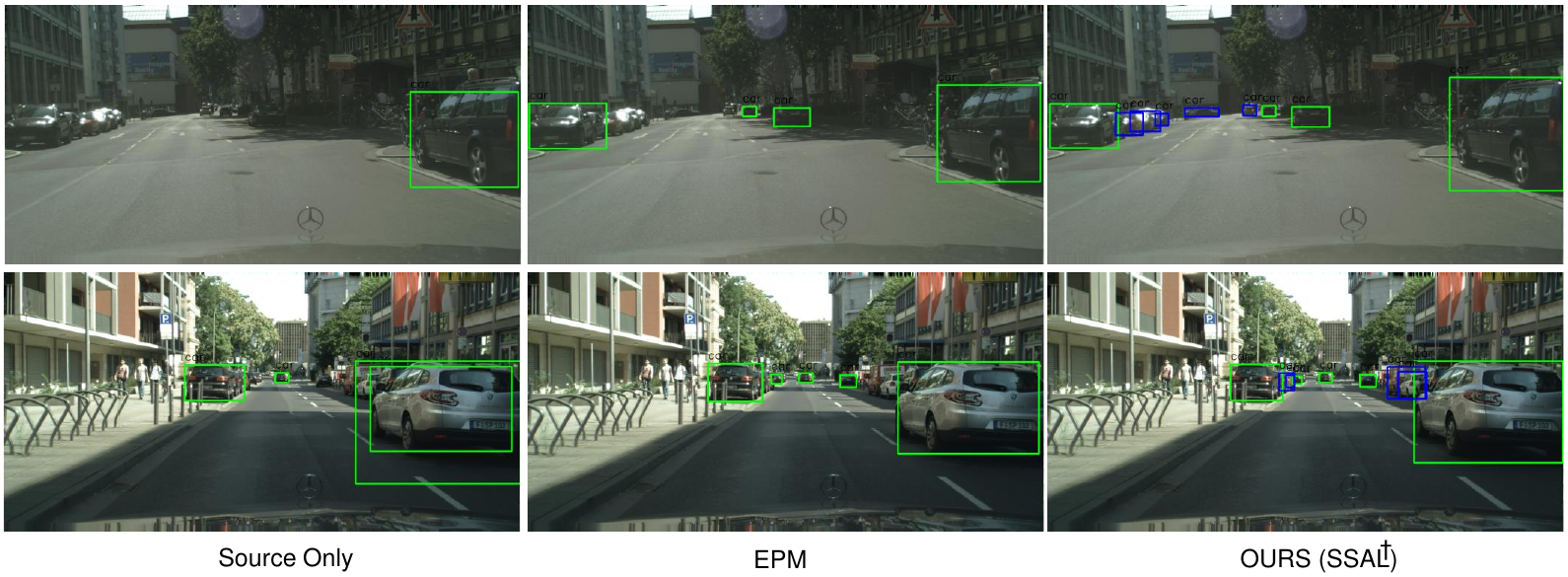}
    \captionsetup{justification=justified}
    \caption{\small \textbf{Sim10k $\rightarrow$ Cityscapes:} ($car$ only) Detections missed by the EPM and found by SSAL\textsuperscript{\textdagger} are shown in \textcolor{black}{Blue}. Compared to EPM~\cite{hsu2020every} our method achieves better adaptation. Better localization in our method results into more accurate detections.}
    %\vspace{-0.3cm}
    \label{fig:qualsim10k}
\end{figure*}

\begin{SCtable}[][t]
    \scriptsize
    \centering
    \renewcommand{\arraystretch}{1.3}
    \tabcolsep=1.9pt\relax
    \begin{tabular}{c|c}
    \hline
        
        Combinations  & AP@0.5 \\ \hline\hline
        UGPL\textsuperscript{\textdagger} & 50.2 \\ \hline
        RandomTiles + UGPL\textsuperscript{\textdagger} & 51.1 \\ \hline
        UGT\textsuperscript{\textdagger} & 51.5 \\ \hline
        Certain Tiles + UGPL\textsuperscript{\textdagger} & 52.2 \\ \hline
        SSAL\textsuperscript{\textdagger} (UGT\textsuperscript{\textdagger}+UGPL\textsuperscript{\textdagger}) & \textbf{53.0} \\ \hline
        
\end{tabular}
\captionsetup{justification=justified}
\caption{\small \textbf{Sim10k $\rightarrow$ Cityscapes}. Comparison of proposed UGT\textsuperscript{\textdagger} vs other tiling strategies, including random and certain tiles in the adversarial learning.}
\label{tab:ugt_impact}
\end{SCtable}

\begin{table}[t]
\scriptsize
\centering
\renewcommand{\arraystretch}{1.3}
\tabcolsep=1.5pt\relax
\begin{tabular}{c|c|c|c|c|c|c|c|c|c|c}
\hline
\textbf{Method}                                       & \textbf{person} & \textbf{rider} & \textbf{car}  & \textbf{truck} & \textbf{bus}  & \textbf{train} & \textbf{mbike} & \textbf{bicycle} & \textbf{\begin{tabular}[c]{@{}c@{}}mAP\\ @0.5\end{tabular}} & \textbf{SO / Gain} \\ \hline\hline
\begin{tabular}[c]{@{}c@{}}\textbf{UGPL}\textsuperscript{\textdagger} \\ w/CT\end{tabular}  & 41.8            & 41.6           & 55.7          & 21.2           & 42.3          & 11.1           & 22.5           & 37.4             & 34.2                                                        & 20.4 / 13.8        \\ \hline
\begin{tabular}[c]{@{}c@{}}\textbf{UGPL}\textsuperscript{\textdagger} \\ w/RT\end{tabular}  & 44.2            & 43.5           & 58.0          & 24.0           & 41.7          & 23.4           & 23.7           & 38.4             & 37.1                                                        & 20.4 / 16.7        \\ \hline
\begin{tabular}[c]{@{}c@{}}\textbf{UGPL}\textsuperscript{\textdagger} \\ w/\textbf{UGT}\textsuperscript{\textdagger}\end{tabular} & \textbf{46.3}   & \textbf{45.8}  & \textbf{59.4} & \textbf{24.8}  & \textbf{45.3} & \textbf{30.6}  & \textbf{26.7}  & \textbf{39.7}    & \textbf{39.8}                                               & 20.4 / \textbf{19.4}        \\ \hline
\end{tabular}
\captionsetup{justification=justified}
\caption{\small \textbf{Cityscapes $\rightarrow$ Foggy Cityscapes}.
Comparison of proposed UGT\textsuperscript{\textdagger} vs other tiling strategies, including random and certain tiles in multi-class adaptation. CT: Certain Tiles, RT: Random Tiles. 
\vspace{-0.8em}}
\label{tab:abtiling}
\end{table}

\begin{table}[t]
\scriptsize
\centering
\renewcommand{\arraystretch}{1.3}
\tabcolsep=1.9pt\relax
\begin{tabular}{c|c|c|c|c|c|c}
\hline 
\textbf{Methods}         & \textbf{AP (mean)}   & \textbf{AP @0.5} & \textbf{AP @0.75} & \textbf{AP @S} & \textbf{AP @M} & \textbf{AP @L} \\ \hline \hline
\text{Source Only}     & 18.1          & 38.0                 & 15.4                  & 4.6                & 21.9               & 37.4               \\ \hline

\text{Baseline}     & 25.9          & 46.0                 & 25.5                  & 5.7                & 28.8               & 52.2               \\ \hline

\text{Confident PL}     & 21.8          & 43.2                 & 19.8                  & 4.7                & 27.5               & 42.9               \\ \hline

Ours (\textbf{UGPL})\cite{munir2021ssal}     & 27.6          & 49.5                 & 26.9                  & 6.7                & 31.2               & 55.0               \\ \hline
Ours (\textbf{UGT})\cite{munir2021ssal}      & 27.5          & 50.0                 & 26.7                  & \text{6.8}       & 31.7               & 54.5               \\ \hline
Ours (\textbf{SSAL})\cite{munir2021ssal} & \text{28.9} & \text{51.8}        & \text{30.4}         & 6.4                & \text{32.7}      & \text{58.7}      \\ \hline

Ours (\textbf{UGPL}\textsuperscript{\textdagger})     & 29.1          & 50.2                 & 29.1                  & 6.0                & 33.5               & \textbf{59.0}               \\ \hline
Ours (\textbf{UGT}\textsuperscript{\textdagger})     & 29.1          & 51.5                 & 29.1                  & \text{6.6}       & 32.7               & 55.5               \\ \hline
Ours (\textbf{SSAL}\textsuperscript{\textdagger}) & \textbf{30.0} & \textbf{53.0}        & \textbf{30.6}         & \textbf{7.0}                & \textbf{35.0}      & \text{58.7}      \\ \hline
\end{tabular}
%\vspace{-0.3cm}
\captionsetup{justification=justified}
\caption{\small \textbf{Sim10k $\rightarrow$ Cityscapes.} Effectiveness of individual components in SSAL and SSAL\textsuperscript{\textdagger}. 
UGPL\textsuperscript{\textdagger} and UGT\textsuperscript{\textdagger} are uncertainty-guided pseudo labels and uncertainty-guided tiling for SSAL\textsuperscript{\textdagger} respectively.
}

\label{tab:abmodules}
\end{table}
% \vspace{-0.5cm}

\begin{table}[t]
\scriptsize
\centering
\renewcommand{\arraystretch}{1.3}
\tabcolsep=1.9pt\relax
\begin{tabular}{c|c|c|c|c|c|c}
\hline 
% \rowcolor[HTML]{33e9ff}
Methods  & \textbf{AP (mean)}   & \textbf{AP @0.5} & \textbf{AP @0.75} & \textbf{AP @S} & \textbf{AP @M} & \textbf{AP @L} \\ \hline \hline
% \rowcolor[HTML]{33e9ff}
SSAL w/o loc & \text{27.2} & \text{48.2}        & \text{26.8}         & \text{5.9}                & \text{31.3}      & \text{54.7}      \\ \hline
% \rowcolor[HTML]{33e9ff}
\textbf{SSAL} & \text{28.9} & \text{51.8}        & \text{30.4}         & \text{6.4}                & \text{32.7}      & \text{58.7}      \\ \hline
% \rowcolor[HTML]{33e9ff}
\textbf{SSAL}\textsuperscript{\textdagger} w/o loc & \text{27.1} & \text{50.3}        & \text{27.1}         & \text{5.9}                & \text{31.0}      & \text{54.9}      \\ \hline
% \rowcolor[HTML]{33e9ff}
\textbf{SSAL}\textsuperscript{\textdagger} & \textbf{30.0} & \textbf{53.0}        & \textbf{30.6}         & \textbf{7.0}                & \textbf{35.0}      & \textbf{58.7}      \\ \hline
\end{tabular}
%\vspace{-0.3cm}
\captionsetup{justification=justified}
\caption{\small \textbf{Sim10k $\rightarrow$ Cityscapes.} Performance after excluding the localization condition for the selection of pseudo-labels and tiles for SSAL and \text{SSAL}\textsuperscript{\textdagger}.}
\label{tab:abwoloc}
\end{table}
% \vspace{-0.5cm}

\noindent \textbf{Weather Adaptation} (Cityscapes $\rightarrow$ Foggy Cityscapes)\textbf{.} 
Under same backbone and detection pipeline, SSAL outperforms the recent one-stage domain adaptive detector (EPM) by an absolute margin of 3.8\% and 1.8\% in terms of mAP and gain. SSAL\textsuperscript{\textdagger} further improves over SSAL by an absolute margin of 0.2\% both in mAP and gain. 
We report (Tab.~\ref{tab:cstofog}) competitive performance against methods built on much stronger, two-stage anchor-based detection pipelines.
In Fig. \ref{fig:qual}, compared to EPM \cite{hsu2020every}, SSAL\textsuperscript{\textdagger} is capable of detecting objects of various sizes under severe climate changes.

\noindent \textbf{Synthetic-to-Real } (Sim10k $\rightarrow$ Cityscapes)\textbf{.} 
SSAL\textsuperscript{\textdagger} delivers a significant gain of 15.0\% (Tab.~\ref{tab:simkittitocs}).
It exceeds the existing state-of-the-art methods, including ones built on stronger detection pipelines and feature backbones, by a notable margin, that is 4.0\% AP over top-performing one-stage adaptive detector (EPM) and $8.1\%$ over two-stage object detection adaptation algorithm SAPNet \cite{li2020spatial}. 
In addition, SSAL\textsuperscript{\textdagger} achieves a considerable improvement of 1.2\% (in AP@0.5) over SSAL\cite{munir2021ssal}.
In Fig. \ref{fig:qualsim10k}, compared to EPM \cite{hsu2020every}, SSAL\textsuperscript{\textdagger} demonstrates better localization accuracy and captures objects of various sizes.

\noindent \textbf{Cross-camera Adaptation} 
(KITTI $\rightarrow$ Cityscapes)\textbf{.} 
For this wide view camera setup to the normal scenario, SSAL achieves $45.6\%$ AP@0.5, compared to results reported by the existing SOTA algorithms using one-stage and two-stage detection pipelines, $43.2\%$ and $42.5\%$, respectively, (Tab.~\ref{tab:simkittitocs}). 
SSAL\textsuperscript{\textdagger} further provides a performance increase of 1.1\% over SSAL \cite{munir2021ssal}.

\noindent \textbf{Large-scale scene Adaptation}
(Cityscapes $\rightarrow$ BDD100k)\textbf{.} 
SSAL\textsuperscript{\textdagger} achieves the best performance among all existing SOTA methods, thereby delivering a significant gain of 8.0\% (Tab. ~\ref{tab:bdd}). It outperforms baseline \cite{hsu2020every} as well as the two-stage detection method \cite{xu2020exploring} by achieving 27.5\% mAP. We see that our method \textcolor{black}{improves over SSAL and} can handle large-scale scene adaptation from a smaller dataset to a larger dataset with complex scenes.

\begin{figure*}[t]
    \centering
    \includegraphics[width=0.7\linewidth]{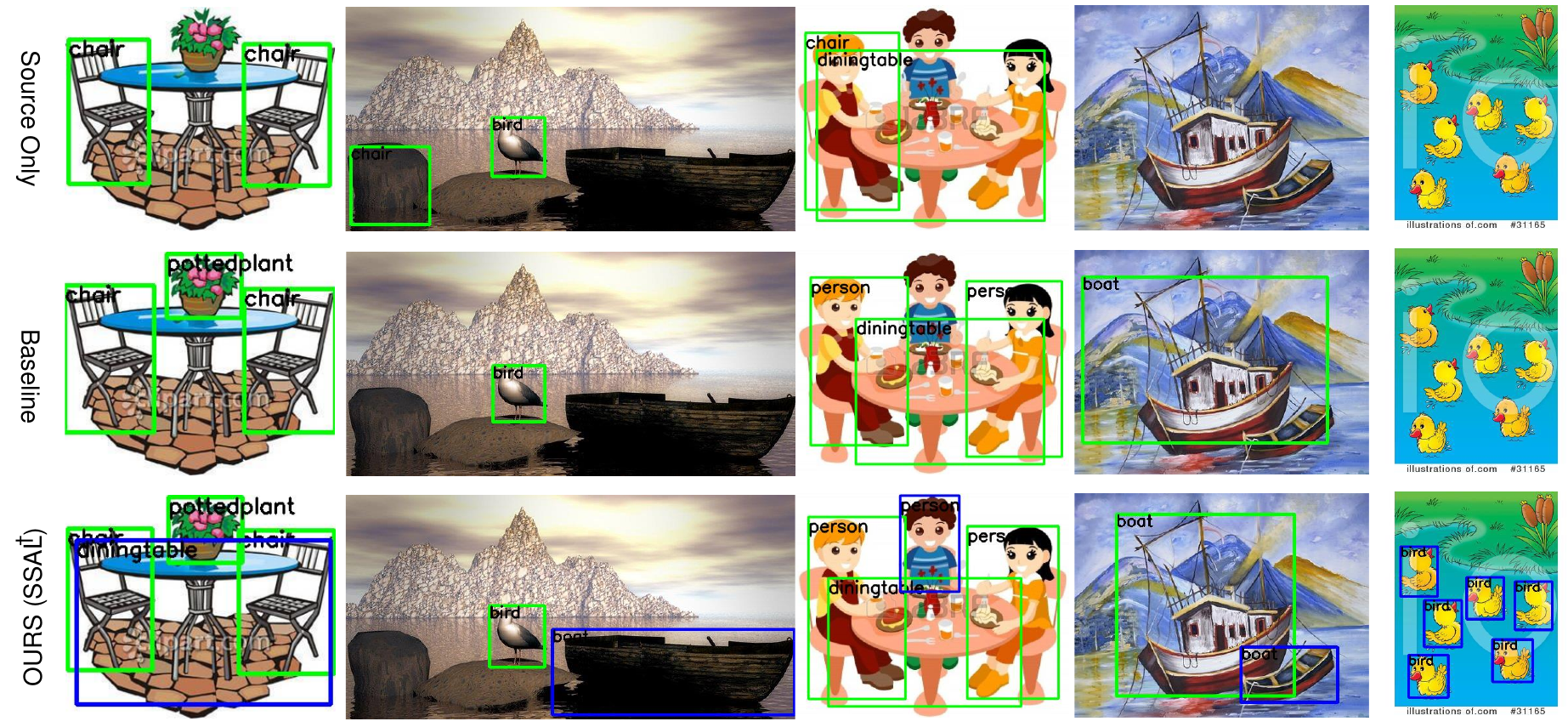}
    \captionsetup{justification=justified}
    \caption{\small Detections missed by the baseline and found by our method (SSAL\textsuperscript{\textdagger}) are shown in \textcolor{black}{Blue}. SSAL\textsuperscript{\textdagger} detects more number of objects under large domain shift compared to EPM \cite{hsu2020every}.\vspace{-0.8em}}
    %\vspace{-0.3cm}
    \label{fig:qualpascal}
\end{figure*}

\noindent \textbf{Severe domain shift}
(PASCALVOC $\rightarrow$ Clipart1k)\textbf{.} 
Under the same backbone (ResNet101) for this scenario, SSAL\textsuperscript{\textdagger} outperforms one- stage domain adaptive detectors
\textcolor{black}{including SSAL}
(Tab. ~\ref{tab:voctoclip}) via achieving 36.5\% mAP with 12.0\% improvement in gain. It provides a 3.0\% \textcolor{black}{\& 2.0\%} improvement compared to existing one-stage domain adaptive methods \textcolor{black}{(WST-BSR \& SSAL respectively).}
In Fig. \ref{fig:qualpascal}, compared to baseline \cite{hsu2020every}, SSAL\textsuperscript{\textdagger} detects more objects under large domain shift .

%\vspace{-0.2cm}
\subsection{Ablation Studies}
\label{sec:Ablation}
%In this section, we study the contribution of individual components, 

\textbf{Contribution of Components:} 
We analyze the effectiveness of individual components in our methods (SSAL and SSAL\textsuperscript{\textdagger}) on \textbf{Sim10k $\rightarrow$ Cityscapes} adaptation (Tab.~\ref{tab:abmodules}).
We first compare the impact on performance by training SSAL with (1) confidence based pseudo-labels only, obtained without our proposed uncertainty based selection, (2) when only uncertainty-guided pseudo-labelling (UGPL) is used without the uncertainty-guided tiling procedure, and (3) when relying only on uncertainty-guided tiling (UGT).
Both UGPL and UGT show an increase of 11.5\% \& 12\% in AP@0.5  over source only model and 3.5\% \& 4.0\%  over our Baseline.
The non-trivial combination of UGPL and UGT, resulting in a synergy between them, produces a further 1.8\% increase in AP@0.5 over their individual performance contributions.
Likewise, we also study the impact of individual component in SSAL\textsuperscript{\textdagger}.
Both UGPL\textsuperscript{\textdagger} and UGT\textsuperscript{\textdagger} show an increase of 12.2\% \& 13.5\% in AP@0.5 over the source only model and 4.2\% \& 5.5\% over the Baseline. Note that, UGPL\textsuperscript{\textdagger} and UGT\textsuperscript{\textdagger} also perform better (in AP@0.5) than their counterparts UGPL and UGT.
SSAL\textsuperscript{\textdagger} further shows 1.5\% \& 2.8\% increase in AP@0.5 over UGPL\textsuperscript{\textdagger} and UGT\textsuperscript{\textdagger}. Notably, it also provide gains over SSAL\cite{munir2021ssal} on a range of IOUs and different object sizes.

\noindent \textbf{UGT\textsuperscript{\textdagger} vs Other Tile Selection Strategies.} We observe the impact of extracting tiles centered around the uncertain detections (UGT\textsuperscript{\textdagger}) for adversarial learning in comparison to different tile selection strategies along with the Uncertainty-Guided Pseudo Labels (UGPL\textsuperscript{\textdagger}) in Tab.~\ref{tab:ugt_impact}. 
Specifically, we chose random tiles, and certain tiles in adversarial learning with UGPL\textsuperscript{\textdagger} in place of proposed tile selection process (UGT\textsuperscript{\textdagger}). Note that, when using random tiles there are various parameters (e.g.,location, size, and aspect ratio) involved in the tile selection process. So, we restrict the tile-selection space using the domain knowledge. Particularly, we restrict that the tile selected should have at least 60\% of the image area. 
In case of \textit{certain tiles}, tiling process is performed around the certain detections for the adversarial learning.
We observe that compared to all three tile selection strategies with UGPL\textsuperscript{\textdagger}, our proposed UGT\textsuperscript{\textdagger} with UGPL\textsuperscript{\textdagger} provides maximum AP@0.5. 
We also compare (UGT\textsuperscript{\textdagger}) vs other tile selection strategies in a more challenging, multi-class adaptation scenario (Tab. \ref{tab:abtiling}). UGT\textsuperscript{\textdagger} provides the best performance, achieving a maximum mAP of 39.8\%, among all other tiles selection strategies.

\noindent \textbf{Impact of object sizes:} In Table~\ref{tab:abmodules}, we also analyze the impact on the performance of different components w.r.t object sizes. In particular, we use MS-COCO evaluation metric \cite{lin2014microsoft} to understand method's behavior with respect to different object sizes categorized as small (S):$<32$ pixels, medium (M): between $32 - 96$ pixels and large (L): $>96$ pixels. 
\noindent \textbf{Impact of localization component:} 
\textcolor{black}{In Table~\ref{tab:abwoloc}, we observe the impact on performance after excluding the localization component in the selection of uncertainty-based pseudo-labels (eq.(\ref{eq:plSelec_3}). We see a notable performance drop in the overall detection performance (1.7\% for SSAL \& 2.9\% for SSAL\textsuperscript{\textdagger} in AP(mean)) and across the full spectrum of object sizes. These results validate the effectiveness of localization component in uncertainty-guided pseudo-label selection.}

\begin{table}[b]
\scriptsize
\centering
\renewcommand{\arraystretch}{1.5}
\tabcolsep=2.9pt\relax
\begin{tabular}{cccc}
\hline
     & \multicolumn{1}{|c|}{R0}       & \multicolumn{1}{c|}{R1}       & R2       \\ \hline
     & \multicolumn{3}{c}{Tiles on Object regions / \% of Total Tiles}                           \\ \hline
SSAL\cite{munir2021ssal} & \multicolumn{1}{|c|}{4955/95.8\%} & \multicolumn{1}{c|}{3262/96.3\%} & 1957/94.9\% \\ \hline
\textbf{SSAL}\textsuperscript{\textdagger} & \multicolumn{1}{|c|}{7002/93.9\%} & \multicolumn{1}{c|}{6546/93.6\%} & 6127/95.0\% \\ \hline
\textbf{$\uparrow$ SSAL $vs$ SSAL\textsuperscript{\textdagger}} & \multicolumn{1}{c}{$\uparrow$ 41.0\%} & \multicolumn{1}{c}{$\uparrow$ 100.7\%} & $\uparrow$ 213.1\% \\ \hline
\end{tabular}
\captionsetup{justification=justified}
\caption{\small We report \# \& \% of tiles out of total tiles (extracted on uncertain regions) capturing object regions over training rounds (R0,R1,R2) for SSAL and SSAL\textsuperscript{\textdagger}. Also, we underline the \%age increase ($\uparrow$) in tiles capturing object regions from SSAL to SSAL\textsuperscript{\textdagger}. \%age increase formula can be seen in supplementary material.\vspace{-0.8em}} 

\label{tab:fppertiles}
\end{table}

\begin{table}[t]
\scriptsize
\centering
\renewcommand{\arraystretch}{1.3}
\tabcolsep=0.3pt\relax
\begin{tabular}{c|c|c|c|c|c|c|c|c} 
\hline
\textit{\textbf{SSAL(PL\textsuperscript{\textdagger})}} & \textit{\textbf{SSAL(FI\textsuperscript{\textdagger})}} & \textit{\textbf{SSAL(ET\textsuperscript{\textdagger})}} & \textbf{AP(mean)} & \textbf{AP@0.5} & \textbf{AP@0.75} & \textbf{AP@S} & \textbf{AP@M} & \textbf{AP@L}  \\ 
\hline
                                        \checkmark      &                                              &                                              & 25.9              & 51.9            & 24.5             & \textbf{8.1}  & 33.7          & 45.3           \\ 
\hline
                                             &      \checkmark                                         &                                              & 26.3              & 49.8            & 25.8             & 7.8           & 34.9          & 48.9           \\ 
\hline
                                             &                                              &   \checkmark                                            & 29.2              & 51.4            & 30.0             & 6.4           & 33.7          & \textbf{59.6}  \\ 
\hline
                                \checkmark              &      \checkmark                                         & \checkmark                                              & \textbf{30.0}     & \textbf{53.0}   & \textbf{30.6}    & 7.0           & \textbf{35.0} & 58.7           \\ 
\hline
\multicolumn{3}{c|}{SSAL\cite{munir2021ssal}}                                                                                                                  & 28.9              & 51.8            & 30.4             & 6.4           & 32.7          & 58.7           \\
\hline
\end{tabular}
%\vspace{-0.3cm}
\captionsetup{justification=justified}
\caption{\small \textbf{Sim10k $\rightarrow$ Cityscapes}. Impact of individual components, specifically introduced in SSAL\textsuperscript{\textdagger}, when they are integrated (either via inclusion or replacement) into the SSAL \cite{munir2021ssal} framework. SSAL(PL\textsuperscript{\textdagger}) denotes SSAL after replacing Eq.(\ref{eq:plSelec}) with Eq.(\ref{eq:plSelec_3}) for pseudo-label selection. SSAL(FI\textsuperscript{\textdagger}) denotes SSAL after including full-sized images, in UGT, for adversarial feature alignment. SSAL(ET\textsuperscript{\textdagger}) is SSAL framework after replacing Eq.(\ref{eq:tileselec}) with Eq.(\ref{eq:tile_1}) for uncertain tile detection.
}
\label{tab:abcomponents}
\end{table}

\begin{table}[t]
\scriptsize
\centering
\renewcommand{\arraystretch}{1.3}
\tabcolsep=0.3pt\relax
\begin{tabular}{c|c|c|c|c|c|c} 
\hline
\textbf{SSAL\textsuperscript{\textdagger}} & \textbf{AP (mean)} & \textbf{AP @0.5} & \textbf{AP @0.75} & \textbf{AP @S} & \textbf{AP @M} & \textbf{AP @L}  \\ 
\hline
% \rowcolor[HTML]{33e9ff}
N=10                 & 30.0               & \textbf{53.0}    & \textbf{30.6}     & 7.0            & 35.0           & 58.7                              \\ 
\hline
% \rowcolor[HTML]{33e9ff}
N=20                 & 30.3               & 52.4             & 30.2              & 7.1            & \textbf{35.8}  & 59.4               \\ 
\hline
% \rowcolor[HTML]{33e9ff}
N=30                 & \textbf{30.5}      & 52.7             & 30.2              & \textbf{7.4}   & 35.6           & \textbf{60.0}  \\
\hline
\end{tabular}
\captionsetup{justification=justified}
\caption{\small \textbf{Sim10k $\rightarrow$ Cityscapes}. Impact of \# of forward passes (N) in SSAL\textsuperscript{\textdagger}. P is the program execution time, N is the forward passes and S is the time consumed for the selection of pseudo labels and generation of tiles.
\vspace{-0.8em}}
\label{tab:Nablation}
\end{table}
%\vspace{-0.1cm}

\begin{table}[b]
    \scriptsize
    \centering
    \renewcommand{\arraystretch}{1.3}
    \tabcolsep=1.9pt\relax
    \begin{tabular}{c|c|c|c}
    \hline
         \textbf{Adaptations (SSAL\textsuperscript{\textdagger})} & \textbf{Source Only} & \textbf{Source + R0} & \textbf{Source+R0+R1+R2} \\ \hline \hline
        CS to Foggy CS & 20.4 & 27.5 & 39.8 \\ \hline
        Sim10k to CS & 38.0 & 47.8 & 53.0 \\ \hline
        KITTI to CS & 34.9 & 40.1 & 46.7 \\ \hline
        CS to BDD100k & 19.5 & 23.8 & 27.5 \\ \hline
        PASCALVOC to Clipart1k & 24.5 & 30.7 & 36.5 \\ \hline
    \end{tabular}
    \captionsetup{justification=justified}
\caption{\small Impact of R0 round. Performing both R1 and R2 rounds (UGPL \textsuperscript{\textdagger} with UGT \textsuperscript{\textdagger}) results in a significant improvement over when only R0 round (UGT\textsuperscript{\textdagger}) is performed.}
\label{tab:impact_R0}
\end{table}

\begin{figure}[!htp]
    \centering
    \includegraphics[width=0.9\linewidth]{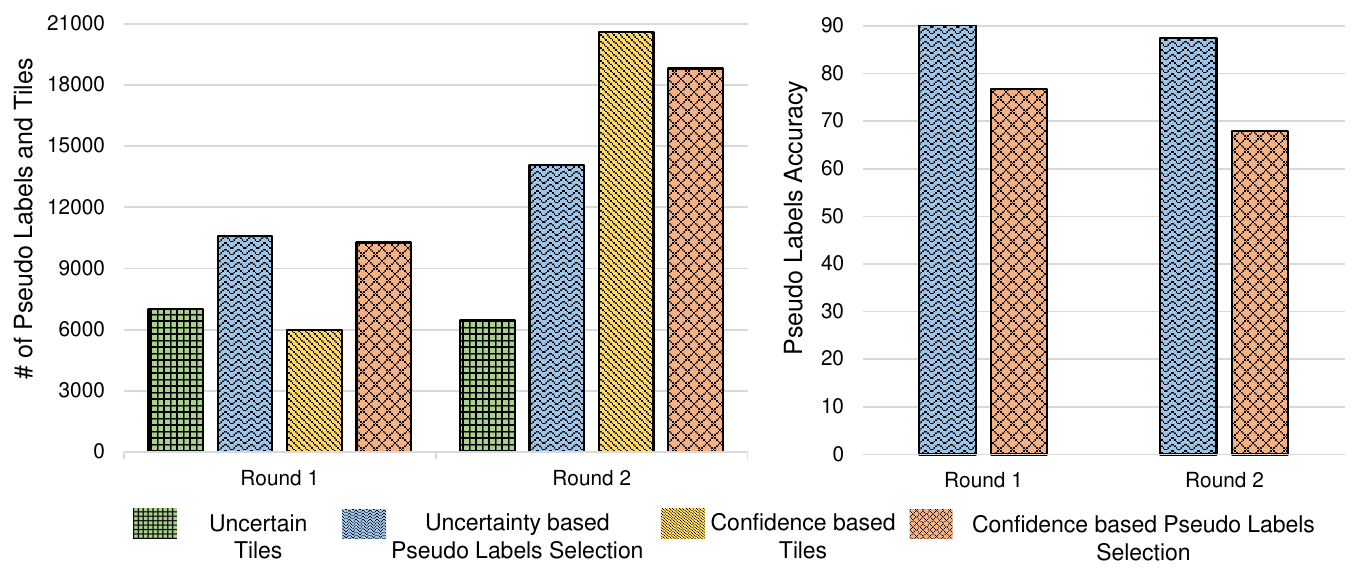}
    \captionsetup{justification=justified}
    \caption{\small 
    \textbf{Left.} Comparison of uncertainty-guided vs the confidence-guided selection of PL and tiles in SSAL\textsuperscript{\textdagger}. 
     \textbf{Right.} Low mean accuracy of confidence-based selected PL vs the uncertainty based PL indicates uncertainty based PL selection is less noisy over the adaptation rounds.  
     As the adaptation process progresses, pseudo-labels (under both types of selection) increases but uncertainty-based PL remains less erroneous than confidence-based PL.\vspace{-0.8em}
    }
    %\vspace{-0.4cm}
    \label{fig:analysis}
\end{figure}

\noindent \textbf{Uncertainty vs Confidence.} 
We contrast between the proposed uncertainty-guided balancing of pseudo-label (PL) selection and the tiling procedure and the confidence-guided balancing of these two procedures (Fig.~\ref{fig:analysis}(left)).
SSAL\textsuperscript{\textdagger} resonates well with the fact that only when the model starts to become more certain of its detections, after round 1, the quantity of selected pseudo-labels should start to increase and so the number of regions being allocated to tiling should begin to decrease. 
This is not the case for the confidence-based balancing. 
Through our adaptive allocation of detection regions, in Fig.~\ref{fig:analysis}(right) we demonstrate that our approach also delivers improved pseudo-labelling accuracy in both the rounds compared to confidence-based selection.

\noindent \textbf{Effectiveness of tiles around uncertain regions.} Tab.~\ref{tab:fppertiles} reports the number/\%age of tiles out of total tiles (extracted on uncertain regions) that capture object regions over the training rounds (R0,R1,R2) both for SSAL and SSAL\textsuperscript{\textdagger}. In general, we see that the uncertainty-guided tiling is capable of capturing object-like salient regions during training evolution. Furthermore, SSAL\textsuperscript{\textdagger} is more effective than SSAL in capturing the same throughout the training rounds.  

\noindent \textbf{Impact of (specific) SSAL\textsuperscript{\textdagger} components.}
We study the impact of individual components, specifically introduced in SSAL\textsuperscript{\textdagger} when they are integrated (either through inclusion or replacement) into the SSAL \cite{munir2021ssal} framework (Tab.~\ref{tab:abcomponents}). SSAL(PL\textsuperscript{\textdagger}) denotes SSAL after replacing Eq.(\ref{eq:plSelec}) with Eq.(\ref{eq:plSelec_3}) for pseudo-label selection. We see that the performance under relatively small and medium objects improve by 1.7\% and 1.0\%, respectively. 
SSAL(FI\textsuperscript{\textdagger}) denotes SSAL after including full-sized images, in UGT, for adversarial feature alignment. We observe a gain of 1.4\% and 2.2\% on small and medium objects, respectively, as FI\textsuperscript{\textdagger} likely facilitates capturing scale variations.
SSAL(ET\textsuperscript{\textdagger}) is SSAL framework after replacing Eq.(\ref{eq:tileselec}) with Eq.(\ref{eq:tile_1}) for uncertain tile detection. It provides a gain of 0.9\% and 1.0\% for large and medium sized objects, respectively. 
Finally, SSAL\textsuperscript{\textdagger}, the combination of SSAL(PL\textsuperscript{\textdagger}), SSAL(FI\textsuperscript{\textdagger}), and SSAL(ET\textsuperscript{\textdagger}), outperforms the respective individual constituents and SSAL in AP(mean), AP@0.5, and AP@0.75 and in medium and large objects.

\noindent \textbf{Impact of increasing forward passes.}
\textcolor{black}{We perform an experiment to study the adaptation performance and time cost as a function of number of forward passes (N). Tab.~\ref{tab:Nablation} reports the results. Upon increasing the value of N from 10 to 20 and 30, we observe slight improvement in AP(mean) of 0.3\% and 0.5\%, in AP of small of 0.1\% and 0.4\%, and in AP of large objects of 0.7\% and 0.6\%. 
}

\noindent \textbf{Impact of R0.} To show how much R0 round contributes to the final performance, we report the performance of the base model (source only) after different rounds of adaptation for all three datasets adaptation scenarios. We report AP@0.5 after R0 and after R0+R1+R2 over the source model. As indicated in Tab.~\ref{tab:impact_R0}, performing both R1 and R2 rounds (that include both UGPL\textsuperscript{\textdagger} + UGT\textsuperscript{\textdagger}) results in significant improvement over when only R0 round (UGT\textsuperscript{\textdagger}) is performed.

\section{Conclusion}
We propose to leverage model's predictive uncertainty to achieve the best of self-training and adversarial learning for domain-adaptive object detection. 
Specifically, we propose to quantify object detection uncertainty by accounting for the variations in the localization prediction and confidence prediction. %\textcolor{blue}{along with its variance spread out} across Monte-Carlo dropout inferences.
Certain detections are considered as pseudo-labels for self-training, while uncertain ones are used to extract tiles (regions in image) for adversarial feature alignment. 
Under various domain shift scenarios, both SSAL and SSAL\textsuperscript{\textdagger} obtains notable improvements over the existing SOTA methods. 

\ifCLASSOPTIONcaptionsoff
  \newpage
\fi

\bibliographystyle{IEEEtran}
\bibliography{natbib}

\vspace{-1.3cm}
\begin{IEEEbiography}[{\includegraphics[width=1.0in,height=1.1in,clip,keepaspectratio]{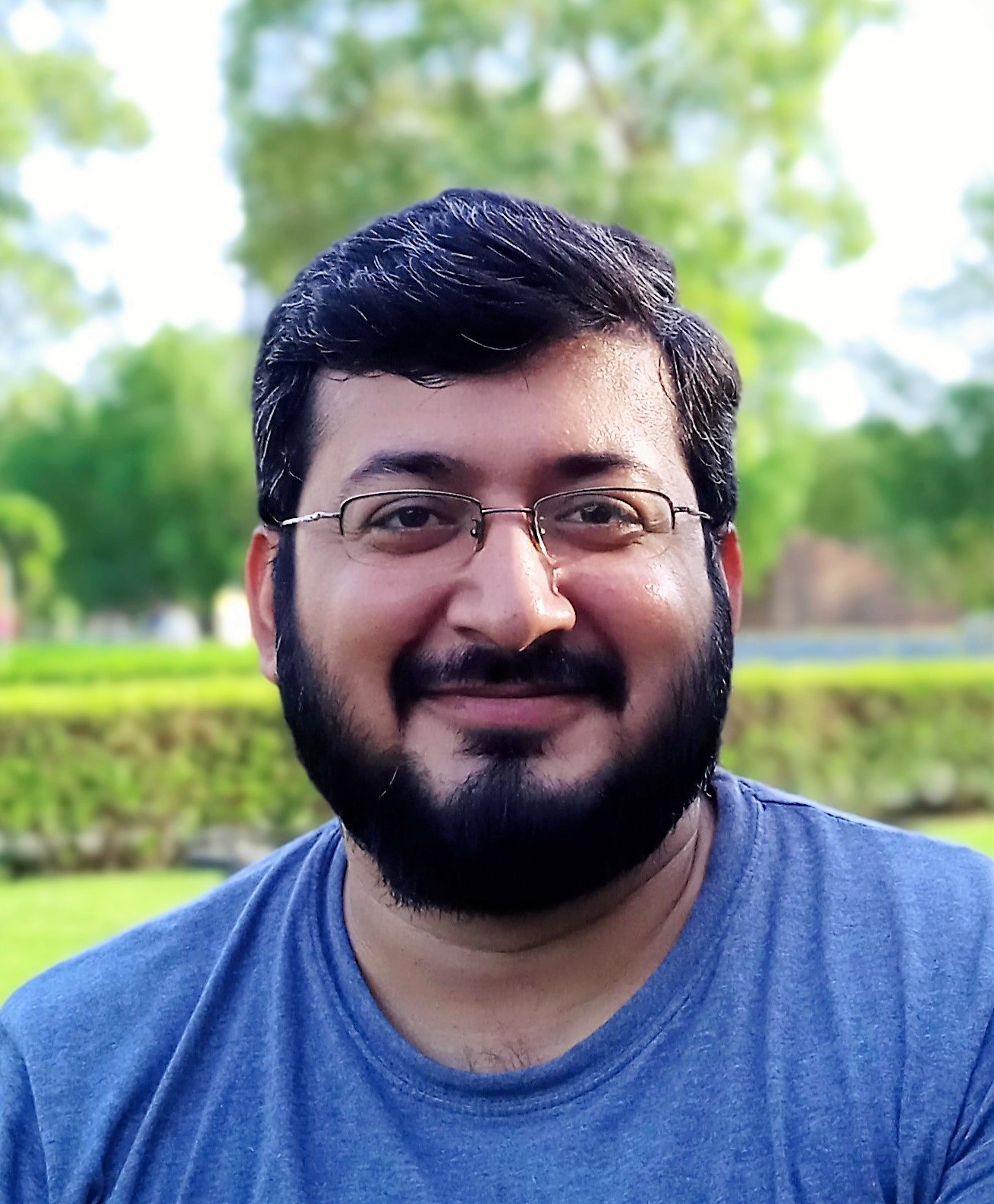}}]{Muhammad Akhtar Munir}
received the BS and MS degrees from COMSATS University, Islamabad, Pakistan. He is currently working toward the Ph.D. degree with Information Technology University, Pakistan, and working as a research associate with Mohamed bin Zayed University of Artificial Intelligence, UAE. He has published papers in reputable machine learning and computer vision venues. His research interests include unsupervised domain adaptation, object detection, and model calibration in deep learning models.
\end{IEEEbiography}
\vspace{-1.3cm}
\begin{IEEEbiography}[{\includegraphics[width=0.9in,height=1.0in,clip,keepaspectratio]{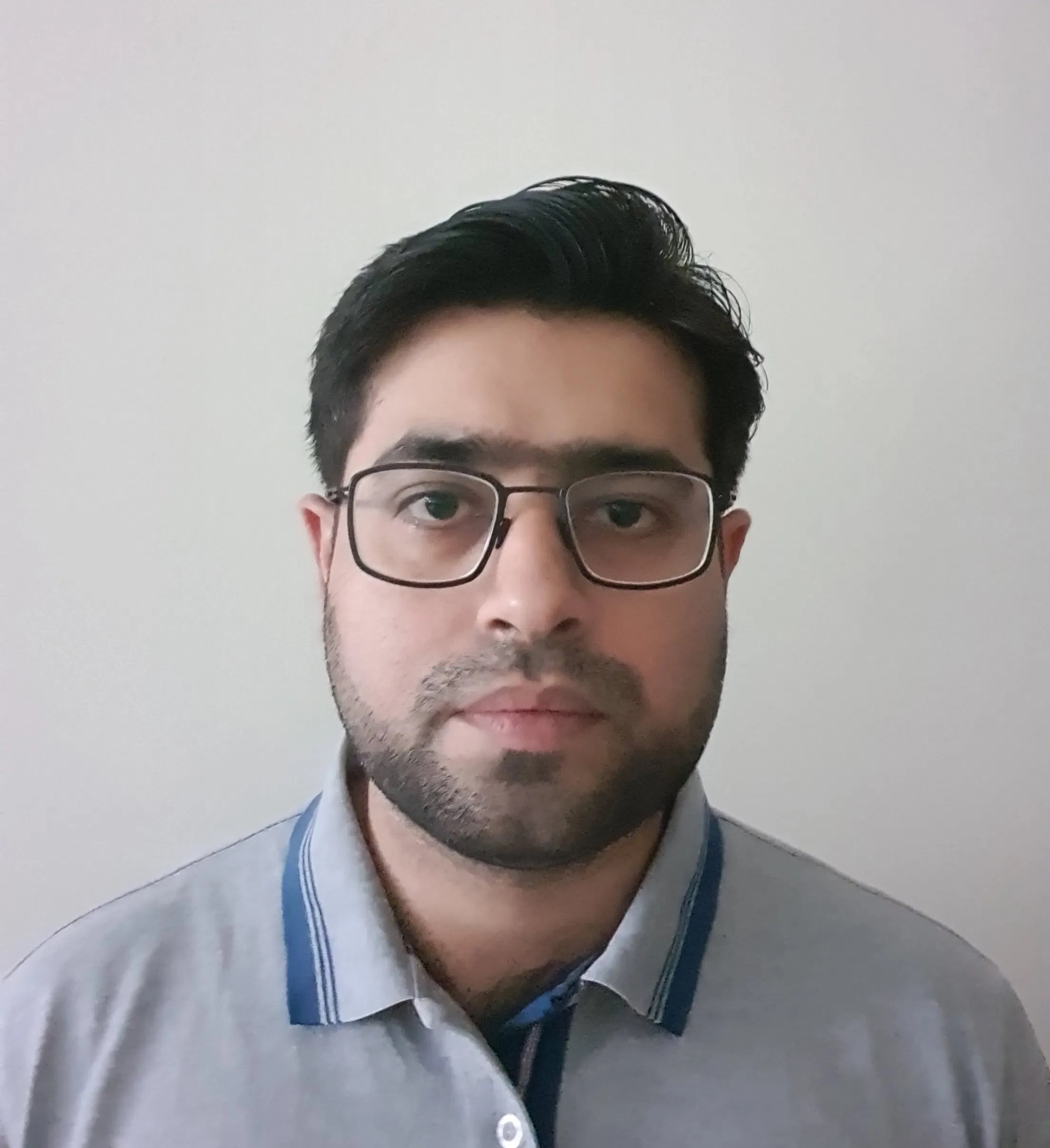}}]{Muhammad Haris Khan} is a faculty member at the Mohamed bin Zayed University of Artificial Intelligence, UAE. Prior to MBZUAI, He was Research Scientist at the Inception Institute of Artificial Intelligence, UAE. He obtained his PhD in Computer Vision from University of Nottingham, UK. He has published several papers in top computer vision venues. His research interests span active topics in computer vision.
\end{IEEEbiography}
\vspace{-1.6cm}
\begin{IEEEbiography}[{\includegraphics[width=1.0in,height=1.0in,clip,keepaspectratio]{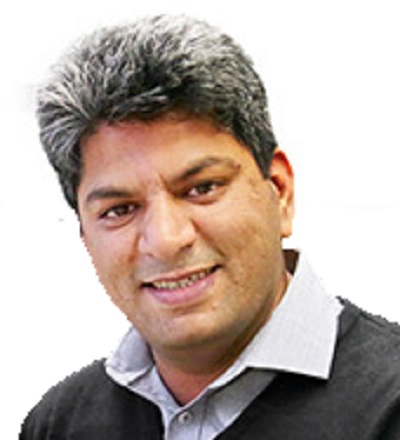}}]{M. Saquib Sarfraz} obtained his PhD in Computer Vision at Technical University
Berlin, Germany in 2009. Currently he works as Lead Deep Learning at Mercedes-Benz Tech Innovation and he also shares his time at Karlsruhe Institute of Technology (KIT) as senior scientist computer vision. He has published several papers in top computer vision venues and have received five best paper awards. His research interests include image \& video understanding, representation learning and clustering.
\end{IEEEbiography}
\vspace{-1.3cm}
\begin{IEEEbiography}[{\includegraphics[width=1.0in,height=1.0in,clip,keepaspectratio]{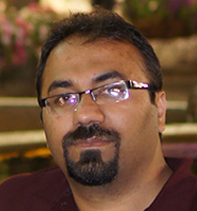}}]{Mohsen Ali}is an Associate Professor at Information Technology University \& a co-founder of the Intelligent Machines Lab. He has been pursuing problem of understanding economic well-being by combining information from satellite imagery and geospatial datasets. His  work has been accepted in respectable computer vision venues. Mohsen obtained a doctorate from the University of Florida. He is a Fulbright alumnus and has been awarded the \textit{Google Research Scholar Award}.

\end{IEEEbiography}

\end{document}